\documentclass{article}
\usepackage{arxiv}
\usepackage{graphicx} %
\usepackage{wrapfig}
\usepackage{amsfonts}
\usepackage{booktabs}
\usepackage{amsthm}
\usepackage{amssymb}
\usepackage{amsmath}
\usepackage{xcolor}
\usepackage{enumitem}
\usepackage{multirow}
\usepackage{float}
\usepackage{tabularx}
\usepackage[letterpaper, left=1.15in, right=1.15in, top=1.1in, bottom=1.1in]{geometry}
\usepackage{parskip}
\usepackage{ragged2e}
\usepackage{hyperref}
\usepackage[style=numeric-comp,sorting=none,natbib=true]{biblatex}
\tbtitle{OMatG-flash: An All-Atom Flow Map with Reinforce Adjoint Matching for Scalable Materials Discovery}
\tbauthors{Thomas Egg\textsuperscript{1}, Harry Winston Sullivan\textsuperscript{2}, Ellad B. Tadmor\textsuperscript{2}, Stefano Martiniani\textsuperscript{1}}
\tbaffil{\textsuperscript{1}New York University, \textsuperscript{2}University of Minnesota}
\tbimage[15mm]{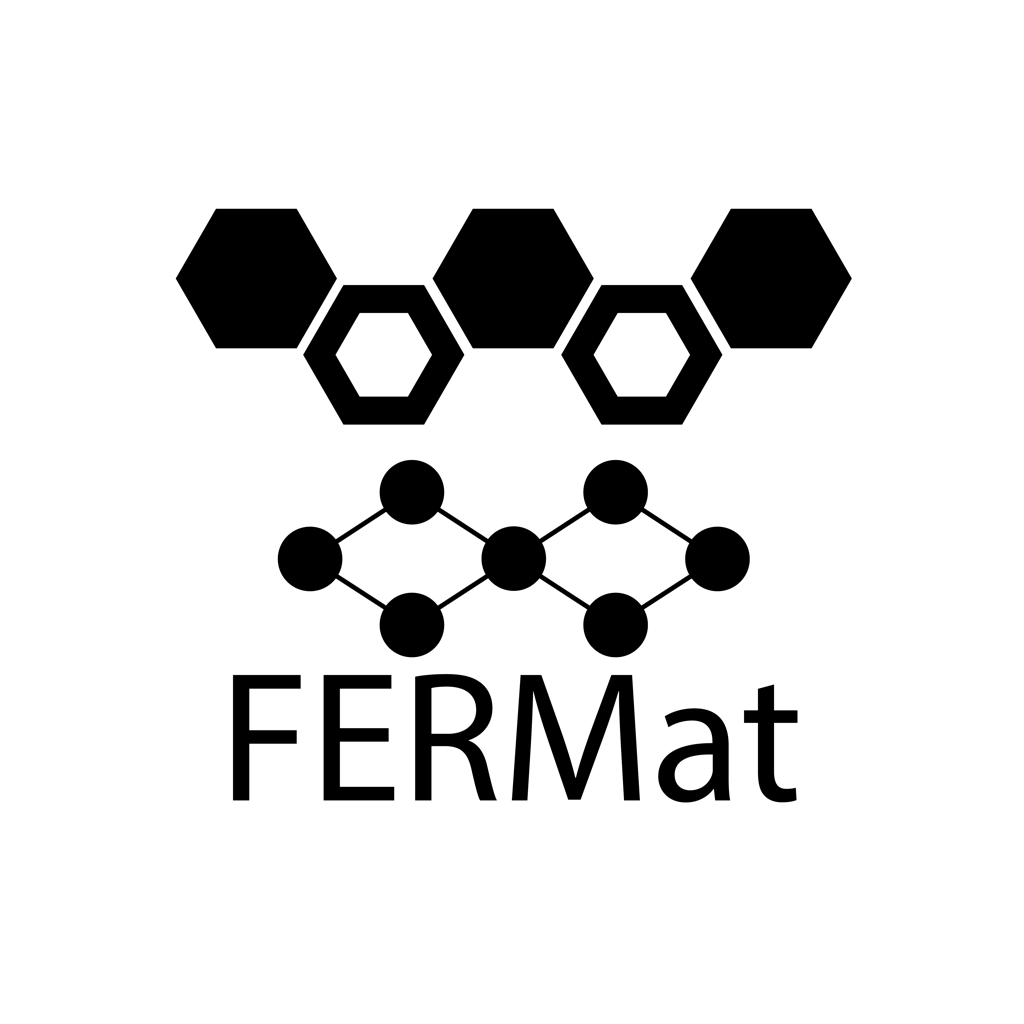}
\tbimagesep{1mm}
\tbimageraise{-2mm}
\tbabstract{\paragraph{Abstract:} The discovery of novel inorganic materials drives technological breakthroughs in critical fields such as computing and energy storage. Generative AI has promised to accelerate the materials discovery pipeline, but state-of-the-art flow and diffusion models remain bottlenecked by the cost of proposing candidate materials.
To address this, we introduce OMatG-flash, an all-atom flow map for inorganic crystal structure prediction (CSP) and de novo generation (DNG). 
OMatG-flash is a Pareto-optimal inference engine for materials, sampling candidate materials with an order of magnitude fewer inference steps and less wall-clock time than existing flow and diffusion models while demonstrating benchmark performance on par with the state-of-the-art.
To enable post-training fine-tuning we apply Reinforce Adjoint Matching to flow maps, further improving match rates and RMSE on the unconditional CSP task. 
OMatG-flash showcases the potential of flow maps to accelerate generation of high-quality candidate inorganic materials and demonstrates a step forward in sample throughput necessary for data-hungry materials discovery workflows.}
\tbmeta{Code}{\url{https://github.com/FERMat-ML/OMatG}}
\begin{document}

    \maketitlebox

\section{Introduction}

A central goal in materials discovery is that of proposing candidate inorganic crystalline materials with desired properties. 
This has implications in critical fields such as computing \cite{chen_high-throughput_2019}, energy storage \cite{pielichowska_phase_2014}, and sustainability \cite{boyd_data-driven_2019}. 
Traditional materials discovery pipelines involve repeated synthesis and experimentation loops which are time-consuming and costly to perform \cite{potyrailo_combinatorial_2011}.
This has inspired the creation of computational algorithms which leverage molecular dynamics and electronic structure to propose materials without the need for lab-based synthesis \cite{de_breuck_generative_2025}.

Such algorithms aim to solve this problem via repeated first-principles quantum chemical calculations which yield accurate results but are computationally expensive and often impractical at scale \cite{pickard_initio_2011}.
Despite unfavorable scaling, the deployment of such methods has culminated in the aggregation of large materials science datasets \citep{Jain2013, zagorac_recent_2019, Cavignac2026Alexandria} along with a wealth of computational tools \citep{ong_python_2013, hjorthlarsen_atomic_2017} for analyzing them.
These tools, coupled with advances in machine learning and artificial intelligence, have spurred the development of highly accurate machine learning models for applications in chemistry and physics which can approximate energies and forces with chemical accuracy \citep{batzner_e3equivariant_2022, batatia2023foundation, wood_uma_2025} or propose candidate compositions and crystal structures \citep{hautier_data_2010, merchant_scaling_2023}.
In the domain of generative modeling for materials, diffusion- and flow-based generative models have shown particularly dominant performance \cite{xie_crystal_2022, jiaoCrystalStructurePrediction2024, miller_flowmm_2024, zeniMatterGenGenerativeModel2024, hoellmer_open_2025a}.
Nevertheless, there is an ever-present need to accelerate the rate at which candidate structures can be proposed. 
As continued effort and innovation flow into scaling autonomous labs \cite{coley_robotic_2019,macleod_selfdriving_2020} and accelerated downstream processing \cite{cohen_torchsim_2025}, so too must the scalability of generative algorithms grow to saturate these systems. 

Flow and diffusion models are fundamentally limited by the need to numerically integrate differential equations to propose candidate materials. 
To address this we introduce OMatG-flash, a transformer-based flow map for CSP and DNG of inorganic materials.
OMatG-flash boasts an order of magnitude faster wall-clock inference time and far fewer neural network evaluations than existing flow and diffusion model architectures while maintaining equivalent performance.
Furthermore, we extend post-training reinforcement learning methods developed for flow matching to flow maps, attaining performance competitive with respect to the state-of-the-art on the CSP task with few function evaluations. 
In contrast to conventional flow map distillation methods, OMatG-flash is trained directly from data unlike teacher-distilled flow maps which require access to a pretrained generative model.
Furthermore, OMatG-flash can be optimized for a target reward via Reinforce Adjoint Matching while retaining its low inference cost.

\paragraph{Our Contributions:} 

\begin{figure}[t]
\centering
\includegraphics[width=0.9\linewidth]{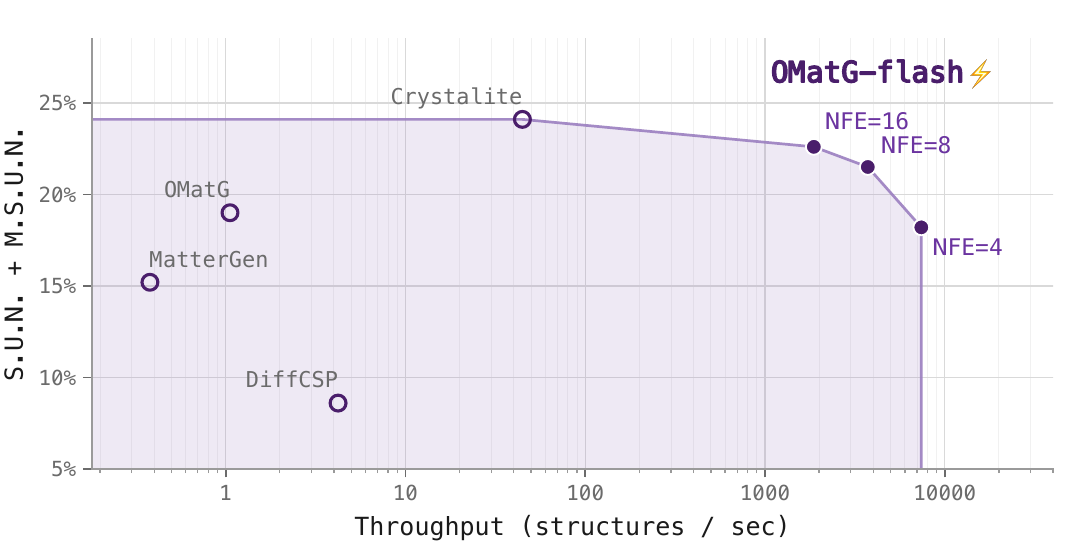}
\caption{\textbf{Pareto-Optimal Inference}. Pareto frontier of combined S.U.N. and M.S.U.N. vs. throughput is shown for existing models. 
OMatG-flash is Pareto-optimal, surpassing existing models by an order of magnitude in terms of throughput while maintaining competitive raw performance on the DNG task. Solid points indicate OMatG-flash under several number of function evaluation (NFE) budgets.}
\label{fig:concept}
\end{figure}

\begin{itemize}[
    leftmargin=*,
    itemsep=1pt,
    topsep=2pt,
    parsep=0pt
]
    \item We introduce OMatG-flash, a Pareto-optimal all-atom flow map that rivals existing flow- and diffusion-based models on materials-generation benchmarks while substantially improving inference-time scalability via few-step sampling (Figure~\ref{fig:concept}).
    \item We show that post-training OMatG-flash with Reinforce Adjoint Matching (RAM)---to our knowledge, its first application to flow maps---achieves state-of-the-art performance on polymorph-aware CSP benchmarks compared to existing diffusion- and flow-based models.
    \item We characterize the key limitations of flow map training for scientific applications and outline prospective future directions for how they may be applied to other problems in materials design.
\end{itemize}
\section{Related Work}
\label{sec:related}
\paragraph{Flows and Diffusions}

Diffusion and flow models formulate generative modeling as a transport problem; mapping a tractable easy-to-sample base distribution to a target distribution that is accessible only through data samples \cite{albergoStochasticInterpolantsUnifying2023}.
The diffusion model approach uses a noising stochastic differential equation (SDE) to construct a regression target for estimating the drift of a reverse denoising SDE. 
With this learned drift, generation proceeds by sampling an initial noisy state and numerically integrating in time to produce an approximate data sample from the target \cite{sohl-dicksteinDeepUnsupervisedLearning2015, songScoreBasedGenerativeModeling2021}. 
In flow models the transport is framed deterministically in terms of an instantaneous velocity field which satisfies the probability flow ordinary differential equation (ODE) describing the probability path of the distribution $\rho_t$ \cite{liu_flow_2022, lipmanFlowMatchingGenerative2023, albergoStochasticInterpolantsUnifying2023}. This velocity field is numerically integrated via deterministic dynamics to draw approximate samples from the target.

\paragraph{Flow Maps and Consistency Models}

Unfortunately, the iterative numerical integration required to draw samples with flow- and diffusion-based models requires repeated model evaluations, hindering time-sensitive performance. 
This has inspired the creation of flow maps which learn a transport that requires few neural network evaluations per generated sample \cite{woo_riemannian_2026, boffiFlowMapMatching2024, boffi_how_2025}.
Flow maps have demonstrated performance matching both diffusion and flow models in various domains at a greatly reduced cost \cite{yoo_selfconditioned_2026, scarpellini_fewstep_2026}.
Flow maps exhibit success in language \cite{lee_flow_2026}, biochemistry \cite{scarpellini_fewstep_2026}, and physics \cite{ripken_learning_2026} while retaining the ability to be learned directly from data.
Furthermore, flow maps are being applied for drawing Boltzmann-distributed data in thermodynamic ensembles \cite{rehman_falcon_2025, ouyang_fewstep_2026}.

\paragraph{Generative Models for Materials Discovery}

Improvement in the performance and scalability of generative AI has inspired the creation of machine learning models which can rapidly propose physically stable and novel crystalline materials.
Diffusion- and flow-based methods have demonstrated particularly strong performance in this domain \cite{xie_crystal_2022, jiaoCrystalStructurePrediction2024, miller_flowmm_2024, zeniMatterGenGenerativeModel2024, zeng_molcrystalflow_2026, lo_fast_2026}.
Subsequent advances in both the underlying algorithms and network architectures further improved generative accuracy \cite{hoellmer_open_2025a, cornet_kinetic_2025, veljković_crystalite_2026}, while conditioning strategies have enabled the generation of materials with targeted properties \cite{kazeev_wyckoff_2025, prakash_guided_2026}.
Current CSP and DNG generative models have thus shifted focus to post-training algorithms with reinforcement learning \cite{park_guiding_2025, hoellmer_open_2026c}, enabling the generative process to push beyond the underlying dataset it was trained on.
Parallel to this, substantial effort has been devoted to training large language models to generate text-based representations of crystal structures \cite{antunes_crystal_2024,xu_plaid_2026, bone_discovery_2026}.

Despite these advances, existing work for inorganic crystals has largely emphasized improvement in sample quality while giving less attention to generation time. 
For large-scale materials screening, the relevant objective is not necessarily the accuracy of any individual prediction, but the number of viable candidates produced within a fixed computational budget.
A model that generates thousands of candidates in the time another requires to generate tens may therefore prove more useful in data-driven materials discovery, even if its per-sample success rate is lower. 
This trade-off is becoming increasingly important as autonomous agents enable high-throughput downstream validation and screening workflows to operate with far less manual intervention \cite{nasri_deterministic_2026}. 
As these workflows scale, generative throughput may become the limiting factor, motivating methods capable of rapidly supplying candidates and saturating downstream screening capacity.

\section{Methods}
\label{sec:methods}

\paragraph{Unit Cell Representation} We represent an $N$-atom periodic unit cell as an element
$\mathbf{c}\in\mathcal{M}$ of the product manifold
\begin{equation}
    \mathbf{c}
    =
    (\mathbf{A},\mathbf{F},\mathbf{y})
    \in
    \mathcal{M}
    :=
    \mathbb{R}^{N\times d_A}
    \times
    \mathbb{T}_{[0,1)}^{N\times3}
    \times
    \mathbb{R}^6,
\end{equation}
where $\mathbf{A}$ is the atomic descriptor matrix with embedding dimension $d_A$, $\mathbf{F}$ represents the
fractional coordinates, and $\mathbf{y}$ is a rotation-invariant unit-cell
representation obtained from the Cholesky decomposition of the lattice metric
tensor $\mathbf{G}=\mathbf{L}\mathbf{L}^\top$ (Appendix~\ref{app:unit_cell}). We define the Riemannian
stochastic interpolant and its endpoint-conditional velocity as
\begin{equation}
    \mathbf{c}_t
    =
    \operatorname{exp}_{\mathbf{c}_0}
    \left(
        \beta_t
        \operatorname{log}_{\mathbf{c}_0}(\mathbf{c}_1)
    \right),
    \qquad
    \dot{\mathbf{c}}_t
    =
    \tfrac{\dot{\beta}_t}{1-\beta_t}
    \operatorname{log}_{\mathbf{c}_t}(\mathbf{c}_1). \label{eq:interpolant}
\end{equation}
The exponential and logarithm maps act componentwise on $\mathcal{M}$
\cite{grenioux_boltzmann_2026}, with their definitions and further details regarding
the crystal representation provided in
Appendix~\ref{app:unit_cell}. We take $\beta_t=t$ to obtain a geodesic interpolant and use a
standard normal base distribution $\mathbf{c}_0\sim\rho_0$, while
$\mathbf{c}_1\sim\rho_1$ is sampled from the dataset.

\paragraph{Flow Matching} The corresponding velocity field which bridges $\rho_0$ and $\rho_1$ is given by a conditional expectation 
$
    \mathbf{v}_t(\mathbf{c})
    =
    \mathbb{E}\!\left[
        \dot{\mathbf{c}}_t
        \mid
        \mathbf{c}_t=\mathbf{c}
    \right].
$
Under suitable regularity conditions, this velocity field generates the
marginal probability path $\rho_t$ induced by the stochastic interpolant
\cite{lipmanFlowMatchingGenerative2023}. An approximation
$\mathbf{v}_t^\theta$ is learned by minimizing
\begin{equation}
    \mathcal{L}_{\mathrm{vel}}(\theta)
    =
    \int_0^1 \mathbb{E}\!\left[
        \left\|
            \mathbf{v}_t^\theta(\mathbf{c}_t)
            -
            \dot{\mathbf{c}}_t
        \right\|^2
    \right] \, \mathrm{d}t
    \label{eq:velocity_matching_objective}
\end{equation}
where the expectation is taken over ${\mathbf{c}}_1\sim\rho_1$ and ${\mathbf{c}}_0\sim\rho_0$. Samples are generated by integrating
\begin{equation}
    \dot{\mathbf{c}}_t
    =
    \mathbf{v}_t^\theta(\mathbf{c}_t),
    \qquad
    \mathbf{c}_0\sim\rho_0,
    \quad t\in[0,1],
    \label{eq:generative_ode}
\end{equation}
giving terminal samples $\mathbf{c}_1\sim\rho_1^\theta\approx\rho_1$ approximating the dataset.

\subsection{Flow Maps}

A two-time flow map $\mathbf{C}_{s,u}$ directly transports samples from the marginal $\rho_s$ to $\rho_u$. For $\mathbf{c}_s\sim\rho_s$, it is defined as the solution to the ODE over the interval $[s,u]$:
\begin{equation}
    \mathbf{C}_{s,u}(\mathbf{c}_s)
    :=
    \mathbf{c}_s
    +
    \int_s^u
    \mathbf{v}_\tau(\mathbf{c}_\tau)\,\mathrm{d}\tau.
    \label{eq:integral_ode}
\end{equation}
Rather than learning only the velocity through
Equation~\eqref{eq:velocity_matching_objective}, a flow map model directly approximates the solution operator
$
\mathbf{C}^\theta_{s,u}\approx\mathbf{C}_{s,u}
$ 
in Equation~\eqref{eq:integral_ode} \cite{song_consistency_2023,boffiFlowMapMatching2024}. 
This enables sampling from $\rho_1$ using a small number of composed function evaluations:
\begin{equation}
    \mathbf{c}_{t_K}
    =
    \left(
        \mathbf{C}^\theta_{t_{K-1},t_K}
        \circ\cdots\circ
        \mathbf{C}^\theta_{t_0,t_1}
    \right)(\mathbf{c}_{t_0}),
    \qquad
    \mathbf{c}_{t_0}\sim\rho_0,
    \qquad
    0=t_0<t_1<\cdots<t_K=1.
    \label{eq:composed_flow_map}
\end{equation}
The terminal sample still satisfies $\mathbf{c}_{t_K}\sim\rho_1^\theta\approx\rho_1$ using only $K$ function evaluations compared to a typical numerical discretization of Equation~\eqref{eq:generative_ode}. In theory $K=1$ is sufficient to draw samples from the target but, in practice, this is often too error-prone.

The flow map must satisfy a \emph{tangency condition}:
\begin{equation}
   \dot{\mathbf{C}}_{t,t}(\mathbf{c}_t)
:=
\left.
\tfrac{\partial}{\partial u}
\mathbf{C}_{t,u}(\mathbf{c}_t)
\right|_{u=t}
=
\mathbf{v}_t(\mathbf{c}_t).
\label{eq:tangent_condition}
\end{equation}
This condition ensures that, in the instantaneous limit, the flow map evolves according to the velocity field that generates the marginal probability path $\rho_t$. 
To enable few-step sampling it must also satisfy a \emph{consistency condition}:
\begin{equation}
    \mathbf{C}_{s,w}(\mathbf{c}_s) =\mathbf{C}_{v,w}\!\left(\mathbf{C}_{s,v}(\mathbf{c}_s) \right), \qquad s<v<w. \label{eq:semigroup_condition}
\end{equation}
This ensures that a composition of steps from time $s$ to $v$, then $v$ to $w$ is equivalent to a single step over the interval $s$ to $w$.

\begin{figure}
    \centering
    \includegraphics[width=\linewidth]{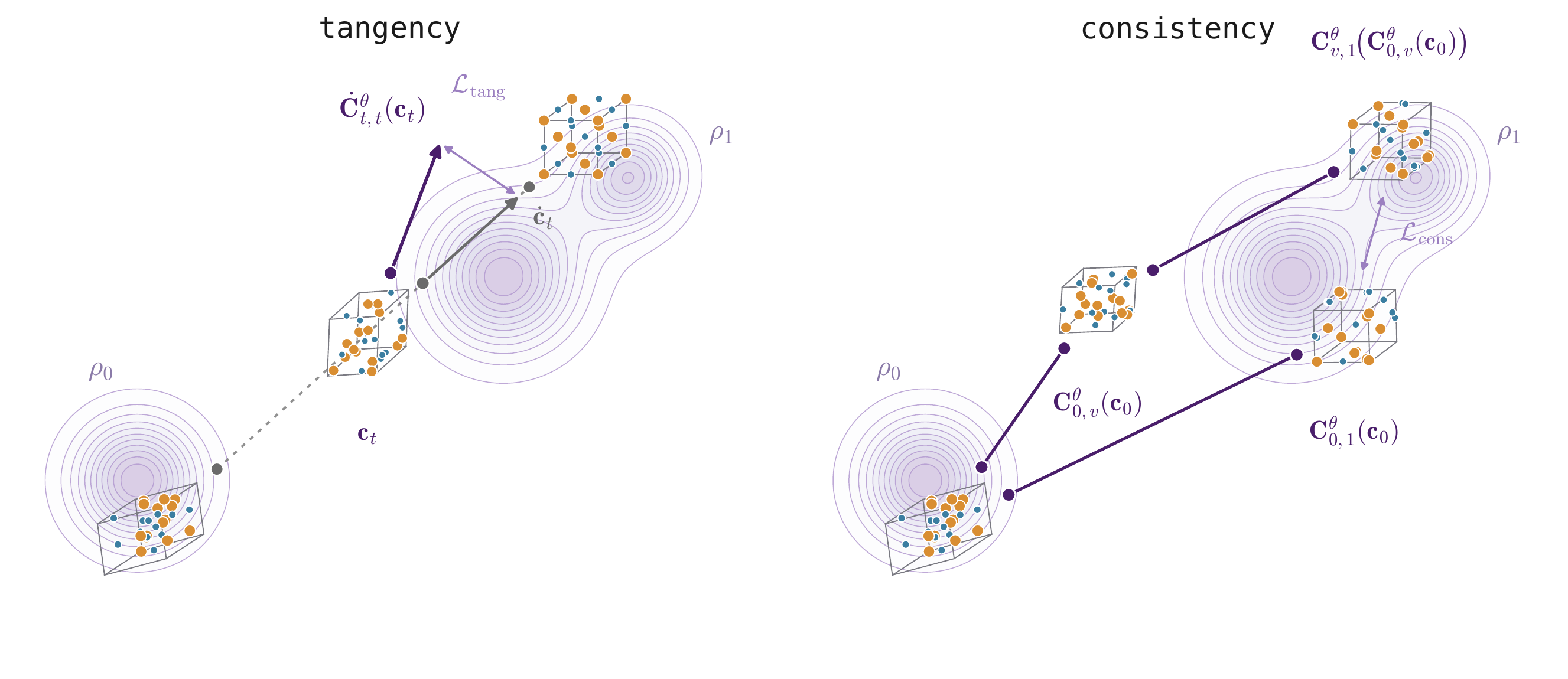}
    \caption{\textbf{Flow Map Training.} OMatG-flash training minimizes a tangency loss (\textbf{Left}) which, similar to standard flow matching, ensures that, at diagonal time $t$, the velocity associated with the instantaneous flow map is aligned with the ground truth velocity field. 
    This is done by drawing an interpolant (dotted gray line) between noise and data and regressing onto the analytical velocity (bold gray arrow) $\dot{\mathbf{c}}_t$. 
    In tandem, a consistency loss (\textbf{Right}) enforces that flow map steps can be properly composed to cover any given time range in a consistent fashion such that few-step sampling properly transports samples from base to target using an average velocity field.}
    \label{fig:flow_map_training}
\end{figure}

\paragraph{Riemannian Flow Maps} We employ a Riemannian MeanFlow \citep{woo_riemannian_2026} to parameterize the flow map jump $\mathbf{C}^\theta_{s, u}$. Specifically, we use an endpoint prediction network $\hat{\mathbf{c}}^\theta_1$ so that the jump from time $s$ to $u>s$ becomes
\begin{equation}
    \mathbf{C}^\theta_{s, u}(\mathbf{c}_s) =  \operatorname{exp}_{\mathbf{c}_s} \left( \frac{u-s}{1-s}   \operatorname{log}_{\mathbf{c}_s} \left( \hat{\mathbf{c}}^\theta_1 (\mathbf{c}_s, s, u) \right)\right).
\label{eq:OMatG-flash_param}
\end{equation}
To compute $\hat{\mathbf{c}}^\theta_1$, we apply Crystalite's neural network architecture \citep{veljković_crystalite_2026}. We combine each atomic descriptor with its corresponding fractional coordinate to form one token per atom and encode the lattice as a separate token, producing $N+1$ tokens in total. 
These tokens are then processed by a two-time scalable interpolant transformer (SiT) backbone, foregoing equivariance for speed \cite{ma_sit_2024,peeblesScalableDiffusionModels2023}. 

A Riemannian flow map \citep{woo_riemannian_2026} is learned by substituting its diagonal velocity $\dot{\mathbf{C}}^\theta_{t,t}$ for the learned velocity in Equation~\eqref{eq:velocity_matching_objective} to enforce tangency. 
We then add a consistency loss \(\mathcal{L}_{\mathrm{cons}}\) to enforce Equation~\eqref{eq:semigroup_condition}. 
The total training objective is finally given by a weighted linear combination of a tangency loss,
\begin{equation}
    \mathcal{L}_\mathrm{tang}(\theta) = \int_0^1 \mathbb{E}\!\left[\gamma_{t} \left\|\dot{\mathbf{C}}^\theta_{t,t}(\mathbf{c}_t)-\dot{\mathbf{c}}_t\right\|^2\right] \, \mathrm{d}t,
\end{equation}
and a consistency loss
\begin{equation}
    \mathcal{L}_\mathrm{cons}(\theta) = \int_0^1 \int_0^w \int_s^w \mathbb{E}\!\left[\frac{\gamma_{s}}{(w-s)^2}\left\|\operatorname{log}_{\mathbf{c}_s}\left(\mathbf{C}^\theta_{s,w}(\mathbf{c}_s)\right)-\texttt{sg}\!\left(\operatorname{log}_{\mathbf{c}_s}\left(\mathbf{C}^\theta_{v,w}\!\left(\mathbf{C}^\theta_{s,v}(\mathbf{c}_s)\right)\right)\right)\right\|^2\right]\, \mathrm{d}v \,  \mathrm{d}s \, \mathrm{d}w.
\end{equation}
A weighted combination of these yields a flow map objective for crystalline unit cells:
\begin{equation}
    \mathcal{L}_\mathrm{total}(\theta) = \lambda_T \mathcal{L}_\mathrm{tang}(\theta) + \lambda_C\mathcal{L}_\mathrm{cons}(\theta).
\label{eq:flow_map_objective}
\end{equation}
Here, $\lambda_{\mathrm{T}}$ and $\lambda_{\mathrm{C}}$ are per-field weights controlling the relative contributions of the tangency and consistency losses, respectively, and $\texttt{sg}$ denotes the stop-gradient operator. $\gamma$ is a time-dependent weighting function chosen for stability \cite{woo_riemannian_2026} (Appendix~\ref{app:generative_transport:flow_map}). The expectations are taken over endpoints $\mathbf{c}_0\sim\rho_0$ and $\mathbf{c}_1\sim\rho_1$ sampled independently. We illustrate flow map training conceptually in Figure~\ref{fig:flow_map_training}.

The above losses can be viewed as self-distillation objectives. However, flow maps can also be trained using a student--teacher framework with a pretrained flow-matching velocity model. In this setting, the velocity predicted by the pretrained model at time $t$ for a noisy sample $\mathbf c_t$ replaces $\dot{\mathbf c}_t$ as the regression target in Equation~\eqref{eq:velocity_matching_objective}.
Our method, OMatG-flash, is \emph{self-distilled} and does not rely on a teacher model, demonstrating that strong flow map performance does not require prior access to a pretrained generative model. For more background on flow maps, see Appendix~\ref{app:generative_transport:flow_map}.

\paragraph{Data Augmentation} OMatG-flash is not translation invariant nor is it rotation equivariant. We learn translation invariance by data augmentation while our canonicalized choice of a cell representation $\mathbf{y}$ obviates the need for data augmentation via randomly sampled rotations as the frame is fixed. More on the architecture of OMatG-flash and the chosen data augmentation strategies are given in Appendix~\ref{app:OMatG-flash_deets}.

\subsection{Reinforce Adjoint Matching}

\begin{figure}
    \centering
    \includegraphics[width=0.95\linewidth]{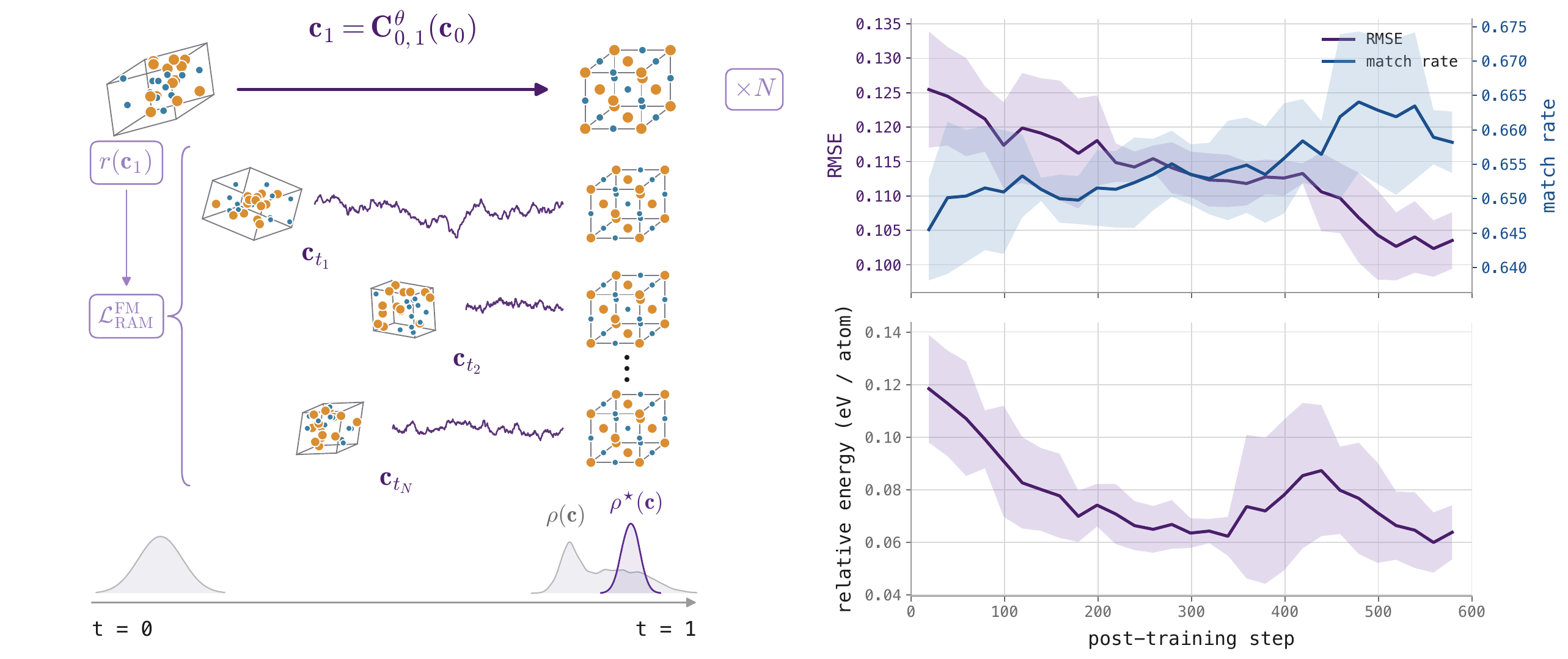}
    \caption{\textbf{OMatG-flash-RAM Post-Training.} A conceptual overview (\textbf{Left}) of RAM post-training for OMatG-flash is presented. OMatG-flash is used to generate a sample, $\mathbf{c}_1$, using few NFE. $N$ replicas of this structure are made and independently renoised. Renoised samples $\{ \mathbf{c}_{t_i} \}_{i=1}^N$ and the reward signal $r(\mathbf{c}_1)$ are used to update the model via the RAM objective in Equation~\eqref{eq:RAM_fm_loss}. (\textbf{Right}) The evolution of relative energy with respect to known crystal structures, RMSE, and match rate all trend desirably during fine-tuning. Relative energies are reported after relaxation, indicating that RAM post-training enables the model to produce structures which minimize to lower energies. Match rate and RMSE are computed before any relaxation.}
    \label{fig:RAMFigure}
\end{figure}

Reinforce Adjoint Matching (RAM) is a method for performing RL to fine-tune flow-based generative models \cite{bergmeister_reinforce_2026}. The objective is to simulate data from a tilted distribution
\begin{equation}
    \rho_1^\star(\mathbf{c}_1) \propto \rho_1(\mathbf{c}_1)e^{r(\mathbf{c}_1)}
\end{equation}
given access only to a scalar-valued reward, $r$, and a reference velocity field, $\mathbf{v}_t^{\mathrm{ref}}$, which can be used to draw samples from $\rho_1$. The core learning task is to estimate the optimally controlled velocity field, $\mathbf{v}_t^\star(\mathbf{c}) = \mathbf{v}_t^{\mathrm{ref}}(\mathbf{c}) + \frac{\sigma_t^2}{2} \nabla_{\mathbf{c}} V_t(\mathbf{c})$ where $\nabla_{\mathbf{c}} V_t(\mathbf{c})$ is the gradient of a value function \cite{domingo-enrich_adjoint_2025, potaptchik_meta_2026}. RAM proposes to learn an approximate optimal control using an expression for this value function gradient minus a kinetic term,
\begin{equation}
\nabla_\mathbf{c} V_t(\mathbf{c})\approx\mathbb{E}\!\left[r(\mathbf{c}_1)\nabla_{\mathbf{c}_t}\log\rho_{1|t}(\mathbf{c}_1\mid\mathbf{c}_t)\,\middle|\,\mathbf{c}_t=\mathbf{c}\right],
\end{equation}
where $\rho_{1|t}$ is the conditional distribution of a clean crystal structure $\mathbf{c}_1$ given a noisy intermediate $\mathbf{c}_t$. This approximation yields a numerically stable and low-variance objective. Under this approximation, we can take a pretrained $\mathbf{v}_t^\theta$ and post-train it by minimizing the following RAM loss to approximate the controlled $\mathbf{v}_t^\star$,
\begin{equation}
    \mathcal{L}_\mathrm{RAM}(\theta) = \int_0^1 \mathbb{E} \left[ \left\| \mathbf{v}^\theta_t(\mathbf{c}_t) - \texttt{sg} \left( \mathbf{v}_t^\mathrm{ref}(\mathbf{c}_t) + r(\mathbf{c}_1) \left( \dot{\mathbf{c}}_t - \mathbf{v}_t^\theta(\mathbf{c}_t) \right) \right) \right\|^2 \right] \, \mathrm{d}t\label{eq:RAM_vel_loss}
\end{equation}
where $\mathbf{v}_t^\mathrm{ref}$ is a frozen copy of $\mathbf{v}_t^\theta$ before any optimization.
Illustrated in Figure~\ref{fig:RAMFigure}, this objective is estimated by generating data, $\mathbf{c}_1$, under the current model $\mathbf{v}_t^\theta$ and noising it analytically via the interpolant in Equation~\eqref{eq:interpolant} to a randomly sampled uniform time $t\in [0,1]$. 
Unlike policy gradient variants applied to diffusion- and flow-based generative models, RAM is formulated to work with \emph{deterministic} inference. 
More background on RAM is given in Appendix~\ref{app:ram}.

\paragraph{RAM for All-Atom Flow Maps}

Flow map inference is not naturally compatible with standard policy-optimization methods such as DDPO \cite{black_training_2024}, PPO \cite{schulman_proximal_2017}, and GRPO \cite{shao_deepseekmath_2024}. Flow maps define deterministic generative dynamics and learn finite-time transport rather than a stochastic policy. Policy-gradient methods require a stochastic process to formulate generation as a Markov decision process.
Existing methods for applying such methods to flow matching navigate around this hurdle by either converting the ODE in Equation~\eqref{eq:generative_ode} to an SDE via the score or by artificially injecting noise into this ODE \cite{liu_flowgrpo_2025, hoellmer_open_2026c}. While these approaches have shown promising results, they do not transfer naturally to flow maps, which do not admit a straightforward score-based reformulation as an SDE except for in the instantaneous limit.
RAM, however, provides a path for fine-tuning a learned flow map.

We adapt RAM to enable flow map reinforcement learning. By exploiting the tangent condition in Equation~\eqref{eq:tangent_condition}, we replace the instantaneous velocity ${\mathbf{v}}^\theta_{t}$ in Equation~\eqref{eq:RAM_vel_loss} with the diagonal derivative of the flow map $\dot{\mathbf{C}}^\theta_{t,t}$. We jointly maintain consistency away from the diagonal via a consistency loss, $\mathcal{L}_{\mathrm{cons}}(\theta)$, to enforce the semigroup property in Equation \eqref{eq:semigroup_condition}. This gives the flow map RAM loss
\begin{equation}
    \mathcal{L}^\mathrm{FM}_\mathrm{RAM}(\theta) = \lambda^\star_\mathrm{RAM}\int_0^1\mathbb{E} \left[ \left\| \dot{\mathbf{C}}^\theta_{t,t}(\mathbf{c}_t) - \texttt{sg} \left( \dot{\mathbf{C}}_{t,t}^\mathrm{ref}(\mathbf{c}_t) + r(\mathbf{c}_1) \left( \dot{\mathbf{c}}_t - \dot{\mathbf{C}}_{t,t}^\theta (\mathbf{c}_t) \right) \right) \right\|^2 \right] \, \mathrm{d}t + \lambda_C^\star\mathcal{L}_{\mathrm{cons}}(\theta)
\label{eq:RAM_fm_loss}
\end{equation}
with chosen weights $\lambda^\star_\mathrm{RAM}$ and $\lambda^\star_C$ indicated with a star to denote post-training. Minimizing this objective requires no reparameterization of the learned flow map nor does it require any sort of noise injection which would corrupt the generative dynamics. 

We implement RAM post-training only for the CSP task using the energy $E$ as a reward model 
\begin{equation}
    r(\mathbf{c}_1) = - \lambda_E E(\mathbf{c}_1).
\label{eq:reward}
\end{equation} 
where $\lambda_E$ is a hyperparameter. This reward pushes the generated distribution towards more stable regions of the energy surface, improving performance at the CSP task. 
We use the MACE-MPA-0 foundational machine learned interatomic potential (MLIP) as the chosen energy model for this task to balance high-fidelity energy evaluation with speed \cite{batatia2023foundation}. 
To stabilize post-training we compute the reward in Equation~\eqref{eq:reward} after taking 50 relaxation steps for each generated structure $\mathbf{c}_1$.
More detail on the OMatG-flash-RAM implementation is provided in Appendix~\ref{app:ram}.

\section{Experiments} \label{sec:experiments}

We train OMatG-flash on two inorganic materials datasets to benchmark performance on CSP and DNG: \textbf{MP-20}, which comprises 45,231 crystal structures curated from the Materials Project \cite{Jain2013}, and \textbf{Alex-MP-20}, which is a larger dataset comprising 675,204 structures from the Materials Project and the Alexandria database \cite{Cavignac2026Alexandria}. 
Each dataset contains crystals with at most 20 atoms in the unit cell and covers a diverse set of compositions and structures. We evaluate OMatG-flash (pretrained) for both CSP and DNG of inorganic materials and OMatG-flash-RAM (post-trained) for CSP.
All reported metrics for competing models are given as reported in existing work \cite{seong_discovering_2026, betala_lematgenbench_2026}.

OMatG-flash is trained on the canonical data splits for both of these datasets \cite{xie_crystal_2022, zeniMatterGenGenerativeModel2024} as well as the new polymorph-split \textbf{MP-20-ps} \cite{martirossyan_all_2025}.
Results with respect to the original data splits are given solely to provide a fair comparison to existing work and we strongly urge future generative modeling efforts to train, evaluate, and compare to the polymorph split (ps) versions of this data using appropriate polymorph-aware metrics as plain MP-20 has known leakage across splits \cite{martirossyan_all_2025}.

\subsection{Crystal Structure Prediction}
A central challenge in inorganic materials design is to predict crystal structure given only the constituent elements of atoms in the unit cell. 
The performance of OMatG-flash at CSP is primarily assessed via match rates and root-mean-squared-error (RMSE) between generated and reference crystals using Pymatgen's \texttt{StructureMatcher} algorithm which is used to report the fraction of generated crystal structures which match an index-paired test-set structure up to a chosen tolerance. 
While informative, match rates depend on arbitrary indexing which is flawed when considering the polymorphism inherent to crystalline materials \cite{martirossyan_all_2025}.
We therefore report and highlight our results on METRe (match-everyone-to-reference) and cRMSE, recently proposed polymorph-aware metrics which improve upon the flawed match rate benchmark. 
METRe addresses the failures of one-to-one match rate by instead searching for a match among \emph{all} generated structures for each test set structure to remove any dependence on arbitrary indexing. 
The associated RMSE values measure the similarity between two matched structures in terms of a unitless distance whereas corrected RMSE (cRMSE) adds a tolerance-based penalty for each non-matching case (Appendix~\ref{app:lemat_metrics}). 
For completeness we report both one-to-one match rates and RMSE along with METRe and cRMSE, but strongly urge wider adoption of polymorph-aware METRe and cRMSE. 
We report the performance of OMatG-flash and OMatG-flash-RAM in Tables~\ref{tab:mr_rmse} and~\ref{tab:metre_mp20ps}. Post-training curves of the match rate, RMSE, and relative energy are shown in Figure~\ref{fig:RAMFigure}.
\begin{table}[t]
\centering
\caption{\textbf{CSP Performance on MP-20 and Alex-MP-20.} OMatG-flash and OMatG-flash-RAM demonstrate strong performance on one-to-one match rate where the subscript indicates the NFE used for inference. Values unavailable in the literature are indicated with dashes. $k$ indicates that inference was performed $k$ times and the best match from each generated set is taken. Each slash-separated entry reports \texttt{valid or invalid} / \texttt{valid}, where the first value is computed over all generated structures and the second only over structures classified as valid. \textbf{Bold} indicates best overall and \underline{underlined} indicates best on a $k=1$ budget.}
\vspace{4pt}
\label{tab:mr_rmse}
\begin{tabular}{l c cc cc}
\toprule
& & \multicolumn{2}{c}{MP-20} & \multicolumn{2}{c}{Alex-MP-20} \\
\cmidrule(lr){3-4} \cmidrule(lr){5-6}
Method & $k$ & MR (\%) $\uparrow$ & RMSE $\downarrow$ & MR (\%) $\uparrow$ & RMSE $\downarrow$ \\
\midrule
DiffCSP     & 1 & 57.82 / 52.51 & 0.0627 / 0.0600 & -- & -- \\
FlowMM      & 1 & 66.22 / 59.98 & 0.0661 / 0.0629 & -- & -- \\
MCFlow      & 1 & \underline{70.38} / 64.08 & \underline{\textbf{0.0592}} / 0.0561 & -- & -- \\
OMatG       & 1 & 69.83 / 63.75 & 0.0741 / 0.0720 & \underline{72.50} / 64.71 & 0.1261 / 0.1251 \\
Crystalite  & 1 & \quad--\quad / \underline{66.09} & \,\quad\,\,--\quad / \underline{\textbf{0.0337}} & \quad--\quad / \underline{68.26} & \,\quad\,\,--\quad / \underline{\textbf{0.0317}} \\
\midrule
\multirow{3}{*}{OMatG-flash$_{16}$}
    & 1 & 67.15 / 60.79 & 0.1172 / 0.1136 & 66.19 / 59.05 & 0.1356 / 0.1342 \\
    & 3 & 75.47 / 68.62 & 0.0954 / 0.0930 & 80.76 / 72.39 & 0.1055 / 0.1042 \\
    & 5 & 78.27 / 71.19 & 0.0878 / 0.0860 & \textbf{85.68} / \textbf{76.91} & 0.0961 / 0.0948 \\
\midrule
\multirow{3}{*}{OMatG-flash$_{16}$-RAM}
    & 1 & 68.90 / 62.55 & 0.0881 / 0.0858 & 70.52 / 62.87 & \underline{0.0955} / 0.0939 \\
    & 3 & 75.76 / 68.89 & 0.0743 / 0.0728 & 81.05 / 72.55 & 0.0807 / 0.0792 \\
    & 5 & \textbf{78.30} / \textbf{71.30} & 0.0713 / 0.0702 & 84.84 / 76.08 & \textbf{0.0764} / 0.0749 \\
\bottomrule
\end{tabular}
\end{table}
\subsection{De Novo Generation}
DNG is the most general materials design problem, requiring proposal of not only the crystal structure but also the composition of elements within a crystal.
We benchmark OMatG-flash's performance at this task by computing stability, uniqueness, and novelty (S.U.N.) and metastability, uniqueness, and novelty (M.S.U.N.) with respect to a known corpus of reference data \cite{siron_lematbulk_2025, betala_lematgenbench_2026}. 
We rely on LeMat-GenBench \cite{betala_lematgenbench_2026} — an open-source leaderboard of generative models for inorganic materials generation —  to compute S.U.N. and M.S.U.N. on 2500 sampled structures in Table~\ref{tab:sun}.
Stability and metastability are determined by comparing the energy of each generated structure to a convex hull of stable phases. 
Uniqueness measures the proportion of generated structures which are distinct chemically or structurally from all other generated structures whereas novelty identifies if a structure is distinct with respect to a chosen reference dataset of inorganic crystals.
To identify matching crystals for uniqueness and novelty we use pymatgen's \texttt{StructureMatcher} algorithm.
More information on the definition and practical calculation of these metrics is given in Appendix~\ref{app:lemat_metrics}. 
We report metrics for models trained on the MP-20 dataset as this dataset has the most reported entries on the LeMat-GenBench leaderboard. 

\section{Discussion}
\subsection{Results}

For both CSP and DNG we compare OMatG-flash to unconditional all-atom diffusion- and flow-based models operating on crystalline unit cells. 
This is done to show how such models for materials can be amortized via a flow map.
There are a variety of models which condition on space group to generate more symmetric structures \cite{levy_symmcd_2025, chang_space_2025, puny_space_2025}. 
Generative models utilizing different representations such as asymmetric units \cite{chang_slayergen_2026} or learned latent spaces \cite{joshi_allatom_2025} have also been shown to yield promising performance for materials.
Lastly, language models and unmasking transformer architectures have also demonstrated success for CSP and DNG \cite{antunes_crystal_2024, kazeev_wyckoff_2025, seong_discovering_2026}. 
MaskGXT \cite{seong_discovering_2026} is of this kind and is the current state-of-the-art overall for CSP on MP-20 and MP-20-ps with benchmark performance that exceeds ours. 
Their inference algorithm performs a double rollout, conditioning the second on the space group representation produced by the first, resulting in improved performance over other CSP models.
For DNG Crystalite \cite{veljković_crystalite_2026} is the current state-of-the-art and is reported as it is an unconditional diffusion model. 
A consensus on which architecture, crystal representation, and choice of conditioning signal is most advantageous for CSP and DNG is still an open question and deserves dedicated study which is out of the scope of this work.

\paragraph{CSP Results}
\begin{wraptable}{r}{0.6\textwidth}
\vspace{-10pt}
\centering
\caption{\textbf{CSP Performance on MP-20-ps.} OMatG-flash-RAM demonstrates state-of-the-art performance on polymorph-aware METRe with respect to existing flow and diffusion models while improving cRMSE. Best is \textbf{bolded} and second-best is \underline{underlined}.}
\label{tab:metre_mp20ps}
\vspace{4pt}
\begin{tabular}{l cc}
\toprule
Method & METRe (\%) $\uparrow$ & cRMSE $\downarrow$ \\
\midrule
DiffCSP              & 53.14 & 0.279 \\
FlowMM               & 65.18 & 0.226 \\
MCFlow               & {70.70} & 0.195 \\
OMatG                & 70.50 & \underline{0.187} \\
Crystalite           & \underline{70.87} & \textbf{0.174} \\
\midrule
OMatG-flash$_{16}$      & 68.56 & 0.250 \\
OMatG-flash$_{16}$-RAM  & \textbf{71.10} & 0.221 \\
\bottomrule
\end{tabular}
\end{wraptable}
Among existing unconditional diffusion- and flow-based generative models OMatG-flash-RAM is state-of-the-art on METRe for MP-20-ps while the base OMatG-flash model is competitive with these methods. 
Post-training OMatG-flash proves remarkably effective for the CSP task overall, simultaneously improving METRe and cRMSE (Table~\ref{tab:metre_mp20ps}) as well as RMSE in every case (Table~\ref{tab:mr_rmse}), demonstrating the effectiveness of the proposed RAM extension and underscoring the potential for flow maps to accelerate high-quality CSP.
While our results for $k>1$ are not meant to be an apples-to-apples comparison with existing results against $k=1$, we argue that our findings emphasize the benefit of rapid, high-quality inference over more expensive inference with slightly improved raw performance.
On MP-20, performing $k$-inference greatly enhances the one-to-one match rates of OMatG-flash.
OMatG-flash enjoys an order of magnitude speedup over existing models. 
Consequently, $k>1$ inference can be done faster than $k=1$ for conventional flow and diffusion models to yield stronger numbers.

Despite high METRe and match rates, the cRMSE and RMSE between ground truth polymorphs and samples produced by OMatG-flash is systematically high for both polymorph-aware and one-to-one matching indicating that, while many structures can be successfully matched to a test set structure, their error relative to that reference is higher than that of a diffusion- or flow-based generative model. 
We argue that this can be attributed to compounding errors in the model exacerbated by performing very few inference steps, a trend that has been observed in few-step inference for physical science applications \cite{woo_riemannian_2026}. 
We also highlight that post-training greatly improves cRMSE and $k$-inference RMSE which suggests that reinforcement with respect to domain-specific rewards can correct for the heightened sensitivity to model error that standard flow maps incur to perform amortized inference. 
Furthermore, assuming that production inference for materials discovery is always followed by MLIP relaxation as in DNG with LeMat-GenBench, we argue that generating structures within the appropriate basin of attraction is of more immediate concern than completely resolving structure.

\paragraph{DNG Results}
\begin{table}[t]
\centering
\caption{\textbf{DNG Performance.} OMatG-flash matches or exceeds all but Crystalite \citep{veljković_crystalite_2026} on S.U.N. (all NFE budgets) and combined S.U.N. and M.S.U.N. (8 or 16 NFE). Novelty is second only to MatterGen for 4 NFE \cite{zeniMatterGenGenerativeModel2024}. Quality of proposed crystals is robust even to very few inference steps. OMatG-flash metrics are computed using LeMat-GenBench code for 2500 structures. Metrics for all other models appear as reported on the leaderboard. Best is \textbf{bolded} and second-best is \underline{underlined}. An asterisk indicates that samples were not prerelaxed before benchmarking according to LeMat-GenBench.}
\vspace{4pt}
\label{tab:sun}
\begin{tabular}{l ccccc}
\toprule
Method & Valid $\uparrow$ & Novel $\uparrow$ & Stable $\uparrow$ & S.U.N. $\uparrow$ & S.U.N. + M.S.U.N. $\uparrow$ \\
\midrule
DiffCSP*     & 95.7 & 66.2 & 2.3 & 0.1  & 8.6  \\
MatterGen   & 95.7 & \textbf{70.5} & 2.0  & 0.2 & 15.2 \\
OMatG       & 96.4 & 51.2 & 11.6 & 1.0 & 19.0 \\
MCFlow      & \underline{97.2} & 52.2 & \underline{11.9} & 0.7 & 19.6 \\
Crystalite  & \textbf{97.2} & 53.2 & \textbf{12.7} & \textbf{1.5} & \textbf{24.1} \\
\midrule
OMatG-flash$_4$ & 96.4 & \underline{66.9} & 7.2 & \underline{1.4} & 18.2 \\
OMatG-flash$_8$ & 96.8 & 63.4 & 7.8 & 1.0 & 21.5 \\
OMatG-flash$_{16}$ & 96.8 & 61.0 & 8.8 & 1.2 & \underline{22.6} \\
\bottomrule
\end{tabular}
\end{table}
OMatG-flash exhibits high-quality performance that remains robust with as few as four model calls per inference. The novelty of structures proposed by OMatG-flash is high on average, trailing only MatterGen~\cite{zeniMatterGenGenerativeModel2024} at 4 NFE while exceeding MatterGen's stability rate by a large margin. The S.U.N. and combined S.U.N. and M.S.U.N. rates of structures generated by OMatG-flash rival the state of the art at a fraction of the inference cost. Except for DiffCSP, competing results given in Table~\ref{tab:sun} are among those which are denoted as prerelaxed on LeMat-GenBench to provide a fair comparison between OMatG-flash and existing models.

We highlight that OMatG-flash-RAM is not benchmarked on this task. As identified in previous work, the S.U.N. and M.S.U.N. metrics are capable of being ``hacked'' by the naive application of RL \cite{chen_accelerating_2025, park_guiding_2025, hoellmer_open_2026c}. 
A model can quickly learn that certain compositions and classes of structures are broadly stable or metastable, focusing generation on repetitive materials to maximize the reward. 
For CSP this is less of an issue, as every proposed material is conditioned on a fixed chemical composition which imposes structure on the energy surface that is difficult to exploit. 
We argue that S.U.N. and M.S.U.N. are more relevant for assessing performance during pretraining but that post-training reinforcement should be targeted to chemical and structural features to target a specific physical property as opposed to stability or novelty in the broadest sense. 
We do not consider RAM post-training for DNG in this work and leave post-training of OMatG-flash and reward design to this end as future work. 

\subsection{Pareto-Optimal Materials Generation}

Flow and diffusion-based models for materials design have demonstrated remarkable capability for sampling stable and novel materials, but the primary advantage of OMatG-flash is scalability at inference time. 
OMatG-flash demonstrates performance competitive with the state-of-the-art but with massively improved inference-time throughput. 
In Figure~\ref{fig:concept} we construct a Pareto frontier of performance vs. inference speed for the DNG task to illustrate this. OMatG-flash exhibits Pareto-optimal performance with respect to existing flow and diffusion models, surpassing all except Crystalite in terms of accuracy on LeMat-GenBench with at least an order of magnitude speedup over existing models. 
Advances in generative AI for science continue to push existing benchmarks further. While improvement in these raw performance indicators is crucial, advancement along other axes such as inference and training speed as well as cost is necessary for scaling high-quality sampling either through hardware or new algorithms. 
OMatG-flash sets the state-of-the-art in balancing rapid inference with high-quality generation.

Conventionally, human judgment and expensive quantum mechanical calculations for property estimation and relaxation have been the bottleneck in materials-discovery as they were often integrated in the proposal engine. As agentic workflows, GPU-accelerated scientific algorithms, and high-accuracy MLIPs are increasingly amortizing and accelerating these downstream filters \citep{ghareeb_multi-agent_2026,nasri_deterministic_2026,seong_discovering_2026, batatia_mace-polar-1_2026}, effort should similarly be focused on amortizing the generation mechanism itself. With this in mind, OMatG-flash is the most capable tool for truly saturating such automated systems and is well-suited to the modern materials science workflow. 

\subsection{Limitations and Future Work}
\label{sec:limitations}

A key limitation of OMatG-flash is the high reported RMSE between generated samples from the base model and ground truth data. 
This, coupled with high METRe rates, indicates that, while OMatG-flash is quite capable of isolating stable crystalline motifs, it struggles to resolve fine-grained structure. 
While post-training notably improves the quality of generated samples in terms of RMSE for the CSP task, improving the quality of the base model is of immediate importance. 
On this note, a possible future direction is to better understand and stabilize RAM post-training for the flow map setting (Appendix~\ref{app:ram}).

An application for OMatG-flash is as an engine for inference time steering where the aim is to take a pretrained generative model and guide it towards desired rewards without changing the weights themselves \cite{wu_practical_2024a, skreta_feynmankac_2025}. 
Recently, there has been increased interest in using flow maps in conjunction with diffusion or flow-based generative models to “look ahead” to $t=1$, compute a reward, and to use the reward or its gradient to bias inference towards regions of feature space with high rewards \cite{potaptchik_meta_2026, holderrieth_diamond_2026, mccallum_strong_2026}. A conditional version of OMatG-flash could be developed which would open the possibility of guidance methods which are grounded in stochastic optimal control \cite{potaptchik_meta_2026, holderrieth_diamond_2026}.

\section{Conclusion}

In this work we introduced OMatG-flash as an all-atom flow map for materials generation and showcased its capability for rapid inference of high-quality inorganic material candidates. Furthermore, we demonstrated post-training capability to facilitate fine-tuning of OMatG-flash based on desired rewards.
Benchmarking on Materials Project data, we show that OMatG-flash is a Pareto-optimal materials generator and pushes the boundaries of inference-time speed while maintaining raw performance on par with the state-of-the-art. 
As materials discovery shifts toward fully automated agentic workflows and accelerated postprocessing, generation time can no longer be treated as a secondary metric.
When one model can generate thousands of candidates in the time another generates one, small differences in per-sample accuracy are overwhelmed by the difference in throughput. 

\section{Code and Data Availability}
The code associated with this work will be released as part of the \href{https://github.com/FERMat-ML/OMatG}{OMatG} package. Relevant model checkpoints will be uploaded to Hugging Face at https://huggingface.co/OMatG. The MP-20 and Alex-MP-20 datasets used for training and evaluating OMatG-flash are all open source.

\section{Acknowledgments}
The authors thank the NYU IT High Performance Computing team for their provision of computational resources and general support. The authors acknowledge funding from NSF Grant OAC-2311632. S. M. acknowledges support from the Simons Center for Computational Physical Chemistry (Simons Foundation grant 839534, MT). The authors gratefully acknowledge use of the research computing resources of the Empire AI Consortium, Inc., with support from the State of New York, the Simons Foundation, and the Secunda Family Foundation.

Large language models assisted with manuscript drafting and editing and research code development; the authors verified all results and claims and take full responsibility for the work.

\newpage

\printbibliography

\newpage

\appendix

\section{Unit Cell Representation}
\label{app:unit_cell}

As introduced in Section~\ref{sec:methods}, we consider the materials proposal problem as inference on crystalline unit cells $\mathbf{c} = (\mathbf{A}, \mathbf{F}, \mathbf{y})$. 
The atomic descriptor matrix $\mathbf{A}$ is obtained by passing the raw atom types $\mathbf{a} \in \mathbb{Z}_+^N$ through a chosen featurization map (Appendix~\ref{app:OMatG-flash_deets}). $\mathbf{F}$ denotes fractional coordinates derived from 3-D Cartesian atomic coordinates $\mathbf{X} \in \mathbb{R}^{N \times 3}$ using the unit cell $\mathbf{L}$
\begin{equation}
    \mathbf{F} = \mathbf{X} \mathbf{L}^{-1}.
\label{cart_to_frac}
\end{equation}
Given any valid cell matrix \(\mathbf{L}' \in \mathrm{GL}^+(3,\mathbb{R})\), we first construct its metric tensor
\[
\mathbf{G}=\mathbf{L}'\mathbf{L}'^{\top}.
\]
We then compute the lower-triangular Cholesky factor \(\mathbf{L}\) such that
\[
\mathbf{G}=\mathbf{L}\mathbf{L}^{\top}.
\]
The corresponding six-dimensional latent representation \(\mathbf{y}\) is then defined by
\[
\mathbf{L}(\mathbf{y}) =
\begin{bmatrix}
e^{y_1} & 0 & 0 \\
y_2 & e^{y_3} & 0 \\
y_4 & y_5 & e^{y_6}
\end{bmatrix},
\]
or equivalently,
\[
\mathbf{y}(\mathbf{L}) =
\left[
\log L_{11},\,
L_{21},\,
\log L_{22},\,
L_{31},\,
L_{32},\,
\log L_{33}
\right] \in \mathbb{R}^6.
\]

\paragraph{Manifold Operations for Unit Cell Transport}
We treat the generation of all components of the crystal unit cell as taking place in continuous space. For $\mathbf{y}$ and $\mathbf{A}$ we operate in Euclidean space with simple Riemannian $\operatorname{exp}$ and $\operatorname{log}$ maps which, for elements $x$ and $y$, are given by
\begin{equation}
    \operatorname{exp}_x(y) = x + y \quad  \mathrm{and} \quad \operatorname{log}_x(y) = y - x.
    \label{eq:explog_euclid}
\end{equation}
For $\mathbf{F}$ we consider generation on the periodic flat torus, $\mathbb{T}_{[0,1)}^{N \times 3} := (\mathbb{R} / \mathbb{Z})^{N \times 3}$ with
\begin{equation}
    \qquad \qquad \qquad \,\, \operatorname{exp}_x(y) = \texttt{wrap} (x + y) \quad  \mathrm{and} \quad \operatorname{log}_x(y) = \texttt{wrap}(y - x + 0.5) - 0.5
\end{equation}
and wrap operation as $\texttt{wrap}(z) = z \,\, \texttt{mod} \,\, 1.0$. These operations allow us to write a Riemannian stochastic interpolant for each component of the crystalline unit cell $\mathbf{c}$.

\section{Generative Modeling as Dynamical Transport}
\label{app:generative_transport}

\subsection{Diffusions}

Diffusion models consider bridging a Gaussian base distribution $\rho_T$ with a target distribution $\rho_0$ by a stochastic process whose time evolution is given by a forward SDE
\begin{equation}
    \mathrm{d} x_t = f_t(x_t) \mathrm{d}t + \sigma_t \mathrm{d} \overrightarrow{W}_t.
\label{eq:diffusion_sde}
\end{equation}
Here $f_t$ is a linear drift $f_t(x) = A_tx+b_t$, $\sigma_t$ is a chosen diffusion coefficient, and $\mathrm{d}\overrightarrow{W}_t$ is an infinitesimal Wiener increment where the arrow indicates that this process moves forward in time. 
This SDE gradually applies noise to data, approaching an isotropic Gaussian as $T$ grows to infinity.
The objective in score-based diffusion models is to learn a reverse-time SDE which generates clean data from noise which, subject to mild conditions on $f_t$ and $\sigma_t$, takes the form \cite{anderson_reversetime_1982, sohl-dicksteinDeepUnsupervisedLearning2015, songScoreBasedGenerativeModeling2021}.
\begin{equation}
	\mathrm{d} x_t = \left[ f_t(x_t) -\sigma_t^2 \nabla_{x_t} \log \rho_t(x_t) \right] \mathrm{d}t + \sigma_t \mathrm{d} \overleftarrow{W}_t.
\label{eq:reverse_diffusion_sde}
\end{equation}
This is generally done by minimizing a score matching objective \cite{hyvärinen_estimation_2005}. If we parameterize a neural network $s_t^\theta$ approximating the score $s_t^\theta \approx \nabla_x \log \rho_t$ then we can write the objective
\begin{equation}
    \mathcal{L}_\mathrm{SM}(\theta) = \int_0^T \mathbb{E} \left[ \left\| s_t^\theta(x_t) - \nabla_{x_t} \log \rho_t(x_t | x_0) \right\|^2 \right] \, \mathrm{d}t
\end{equation}
Since the forward SDE has an affine-linear drift and a state-independent diffusion coefficient, its transition kernel $\rho_t(x_t | x_0)$ is Gaussian and given by the usual solution to the SDE in Equation~\eqref{eq:diffusion_sde} \cite{uhlenbeck_theory_1930}.
With access to an estimator for the score, Equation~\eqref{eq:reverse_diffusion_sde} can be integrated backwards in time to generate data from independent realizations of $\rho_T$.

\subsection{Flows}

Following the success of diffusion models, methods have been formulated to learn a deterministic transport between noise and clean data. 
In this flow-based generative modeling framework a coupling between a base distribution $\rho_0$ and a target $\rho_1$ is given by a velocity field \cite{lipmanFlowMatchingGenerative2023, albergoStochasticInterpolantsUnifying2023, liu_flow_2022}.\footnote{Literature on flows and diffusions is often inconsistent in terms of which direction in time points from noise to data. In diffusion model literature, noise is also attributed to a steady-state realized at $T=\infty$ whereas $T=0$ is a clean initial condition that is progressively noised. In flow-based models, $t=0$ often corresponds to noise and $t=1$ to data. We choose this convention throughout the paper as our method is more directly related to flow models.} 
The typical approach is to bridge noise and data with an interpolant as in Section~\ref{sec:methods},
\begin{equation}
    x_t = \alpha_t x_0 + \beta_t x_1,
\label{eq:interpolant_app}
\end{equation}
where $x_1$ is sampled from the target and $x_0$ is sampled from noise. 
Both $\alpha_t$ and $\beta_t$ are scheduling functions which determine how data is noised in time. 

Interpolation between noise and data allows for the construction of the flow matching algorithm which aims to learn an instantaneous velocity field $v_t$.
The desired velocity is that which solves a transport equation
\begin{equation}
    \frac{\partial \rho_t}{\partial t} + \nabla \cdot (v_t \rho_t) = 0.
\label{eq:ode_transport}
\end{equation}
This transport equation describes the evolution of a time-dependent density from $t=0$ to $t=1$. 
It can be shown that this velocity can be written in terms of the interpolant in Equation~\eqref{eq:interpolant_app}
\begin{equation}
    v_t(x) = \mathbb{E} [\dot{x}_t | x_t = x]
\end{equation}
We approximate this velocity with a neural network $v_t^\theta$ by minimizing the conditional flow matching loss
\begin{equation}
    \mathcal{L}_\mathrm{CFM}(\theta) = \int_0^1 \mathbb{E} \left[ \left\| v^\theta_t(x_t) - \dot{x}_t \right\|^2 \right] \, \mathrm{d}t
\end{equation}
where the expectation is taken over $(x_0,x_1)$ pairs. Access to an approximate $v_t$ allows for the generation of new data samples through integration of an ODE forwards in time

\begin{equation}
    \frac{\mathrm{d} x}{\mathrm{d}t} = v_t, \quad \mathrm{s.t.} \quad x_0 \sim \rho_0.
\label{eq:cfm_ode}
\end{equation}

Flows and diffusions are closely related paradigms for generative modeling \cite{albergoStochasticInterpolantsUnifying2023}. If $\rho_0$ is a standard Gaussian, it is possible to simulate data via a diffusion SDE similar to Equation~\eqref{eq:reverse_diffusion_sde} given access to a flow velocity field $v_t$ using the following relation
\begin{equation}
    \nabla_x \log \rho_t(x) = \alpha_t^{-1} \frac{\beta_t v_t(x) - \dot{\beta}_t x}{\dot{\beta}_t \alpha_t - \beta_t \dot{\alpha}_t }.
\label{eq:reparam}
\end{equation}
This relationship is used for deriving Flow-GRPO \cite{liu_flowgrpo_2025}. The relationship between the score and the velocity is similarly used in the derivation of RAM (Appendix~\ref{app:ram}).

\subsection{Flow Maps}
\label{app:generative_transport:flow_map}

The success of flows and diffusions has inspired exploration into flow maps and consistency models \cite{song_consistency_2023, boffiFlowMapMatching2024, boffi_how_2025}. Given a base and target distribution bridged by Equation~\eqref{eq:ode_transport} a flow map is a two-time map $X_{s,u}$ which satisfies

\begin{equation}
    X_{s,u}(x_s) = x_u
\end{equation}

where $s,u \in [0,1]$. 

A flow map must satisfy a tangent and consistency condition. The tangent condition enforces that, in the instantaneous limit, the flow map velocity matches the velocity in Equation~\eqref{eq:cfm_ode}

\begin{equation}
\dot{X}_{s,s}(x_s)
:=
\left.\frac{\partial}{\partial u}X_{s,u}(x_s)\right|_{u=s}
=
v_s(x_s).
\label{eq:tangent_condition_limit}
\end{equation}

A flow map must also be consistent through time. There are three conditions that can be used to enforce consistency:

\begin{itemize}
    \item \textbf{Lagrangian}
    \begin{equation}
        \frac{\partial X_{s,u}(x_s)}{\partial u} = v_{u}(X_{{s,u}}(x_s))
    \end{equation}

    \item \textbf{Eulerian}
    \begin{equation}
        \frac{\partial X_{s,u}(x_s)}{\partial s} + \nabla X_{{s,u}}(x_s) v_s(x_s) = 0
    \end{equation}

    \item \textbf{Semigroup}
    \begin{equation}
        X_{{v,w}}(X_{s,v}(x_s)) = X_{s,w}(x_s)
    \end{equation}
    
\end{itemize}

where $s<v<w$. Flow maps have been successfully generalized for generative modeling on Riemannian manifolds \cite{davis_generalised_2026, woo_riemannian_2026}. As outlined in Section~\ref{sec:methods} we employ Riemannian MeanFlow with a semigroup consistency objective for training OMatG-flash. Following this prescription \citep{woo_riemannian_2026}, we parameterize OMatG-flash as a two-time map
\begin{equation}
    X_{s,u}(x_s) = \exp_{x_s} \left( ( u-s ) v_{s,u}(x_s)\right)
\end{equation}
where $\exp_xy$ is the exponential map associated with a manifold $\mathcal{M}$ and $s<u$. We find strong performance by parameterizing the average velocity field $v_{s,u}$ with endpoint prediction. Specifically, given $x_s$ and an estimator of the endpoint $x_1(x_s, s, u)$ we write the two-time velocity
\begin{equation}
    v_{s,u}(x_s) = \frac{\log_{x_s} x_1(x_s, s, u)}{1-s}
\end{equation}
where $\log_xy$ is the logarithm map on $\mathcal{M}$. Composing these two equations above we arrive at the parameterization of OMatG-flash in Equation~\eqref{eq:OMatG-flash_param}. For both tangency and consistency losses we employ weighting functions to stabilize the loss $\gamma_{t} = \frac{1-t}{\max(1-t, \epsilon)}$. We set $\epsilon=0.1$.

\section{OMatG-flash Architecture}
\label{app:OMatG-flash_deets}

\paragraph{Atom Type Encoding} 
OMatG-flash applies Crystalite's subatomic tokenization scheme \citep{veljković_crystalite_2026}. For atomic number $Z_i$, Crystalite constructs the 34-dimensional descriptor
\begin{equation}
\mathbf{d}(Z_i)
=
\left[
\mathbf{e}_{r(Z_i)}^{(7)}
\,\middle\|\,
\mathbf{e}_{g(Z_i)}^{(19)}
\,\middle\|\,
\mathbf{e}_{b(Z_i)}^{(4)}
\,\middle\|\,
\left(
\tfrac{n_s}{2},
\tfrac{n_p}{6},
\tfrac{n_d}{10},
\tfrac{n_f}{14}
\right)
\right],
\end{equation}
where $\mathbf{e}^{(k)}$ denotes a $k$-dimensional one-hot encoding, $r$, $g$, and $b$ denote the element's period, group, and block, and $(n_s,n_p,n_d,n_f)$ are its ground-state valence-shell occupancies. 
After standardization, group weighting, PCA projection, and normalization, each descriptor is mapped to $\mathbf{a}(Z_i)\in\mathbb{R}^{d_A}$ with $d_A=24$. The atomic descriptor matrix is formed by stacking these vectors:
\begin{equation}
\mathbf{A}
=
\begin{bmatrix}
\mathbf{a}(Z_1)^\top\\
\vdots\\
\mathbf{a}(Z_N)^\top
\end{bmatrix}
\in\mathbb{R}^{N\times d_A}.
\end{equation}
This representation effectively describes the space of elements and circumvents the need for a discrete diffusion because the state is part of a continuous manifold.

\paragraph{Atom Type Decoding}
After integration is complete, each generated atomic descriptor is snapped to the closest valid element descriptor in PCA space. Specifically, for a generated descriptor $\hat{\mathbf{A}}_{i,1}$, we assign the atomic number
\begin{equation}
    \hat{Z}_i
    =
    \operatorname*{arg\,max}_{Z\in\{1,\ldots,100\}}
    \frac{
        \left\langle \hat{\mathbf{A}}_{i,1},\mathbf{a}(Z)\right\rangle
    }{
        \left\|\hat{\mathbf{A}}_{i,1}\right\|_2
        \left\|\mathbf{a}(Z)\right\|_2
    }.
\end{equation}
Decoding is just a nearest-neighbor projection under cosine similarity to the set of valid PCA-projected element descriptors.

\paragraph{Neural Network Architecture}
We parameterize OMatG-flash as an all-atom transformer, again building off of the Crystalite architecture to do so \cite{veljković_crystalite_2026}. 

The atoms in a unit cell $\mathbf{c}$ are tokenized into $N$ tokens, one per atom, with one additional token for the lattice. The tokenization of an atom indexed $i$ is a linear combination of a coordinate and atom type embedding
\begin{equation}
    \mathbf{h}^{\mathrm{atom}}_{i} = E_{\mathrm{comp}}(\mathbf{A}_i) + E_{\mathrm{coord}}(\mathrm{SinEmb}(\mathbf{F}_i; N_{\mathrm{freq}}^{\mathrm{coord}}))
\end{equation}
where $\mathrm{SinEmb}$ featurizes the coordinates via a Fourier expansion,
\begin{equation}
    \mathrm{SinEmb}(x; N_\mathrm{freq}) = [\sin(2\pi kx), \cos(2\pi kx)]_{k=1}^{N_\mathrm{freq}},
\end{equation}
and $E_\mathrm{comp}, E_{\mathrm{coord}}$ are two-layer MLPs with a SiLU activation. The lattice token is similarly obtained by passing $\mathbf{y}$ through a two-layer MLP embedding layer
\begin{equation}
    \mathbf{h}^{\mathrm{lattice}}_{N+1}(\mathbf{y}) = E_\mathrm{lattice} (\mathbf{y})
\end{equation}
yielding a sequence of tokens $\mathbf{t}^{(0)} =
[\mathbf{h}_1^\mathrm{atom},\ldots,\mathbf{h}_N^\mathrm{atom},
\mathbf{h}_{N+1}^\mathrm{lattice}]$ which the all-atom transformer processes through layers of self-attention. The superscript indicates that this is the initial tokenized embedding before any transformer layers are applied. OMatG-flash is adapted from the diffusion transformer architecture \cite{peeblesScalableDiffusionModels2023}. Our conditioning vector is derived using the left and right foot times. For a source time $s$ and a destination time $u$ this conditioning vector is predicted via a two-layer MLP on $s$ and $\Delta = u - s$
\begin{equation}
    \mathbf{h}_{\mathrm{conditioning}}(s, u) = E_{\mathrm{cond}}\left(\mathrm{SinEmb}(s;N_\mathrm{freq}^\mathrm{time}), \mathrm{SinEmb}(\Delta;N_\mathrm{freq}^\mathrm{time})\right)
\end{equation}
$\mathbf{h}_\mathrm{conditioning}$ along with $\mathbf{t}^{(0)}$ are passed into the model. $\mathbf{h}_\mathrm{conditioning}$ is used to predict scale and shift parameters which normalize the tokens via adaptive layer normalization (adaLN) over the course of several transformer layers. 

We enable physics-informed self-attention within the model using Crystalite's geometry enhancement module. This module computes a block-diagonal bias mask, $\mathbf{B}(\mathbf{c})$, that modulates attention only between atom tokens $i$ and $j$ by first computing the approximate minimum-image displacement using $\Delta \mathbf{f}_{ij}(\mathbf{r}) = \mathbf{f}_j - \mathbf{f}_i + \mathbf{r}$,
\begin{equation}
    \mathbf{r}^\star_{ij} = \arg \min_{\mathbf{r} \in \{-1, 0, 1\}^3} \Delta \mathbf{f}_{ij}(\mathbf{r}) \mathbf{G}(\mathbf{y}) \Delta \mathbf{f}_{ij}^\top(\mathbf{r}),
\end{equation}
then 
\begin{equation}
    \Delta\mathbf{f}_{ij}^\mathrm{mi} = \mathbf{f}_j - \mathbf{f}_i + \mathbf{r}_{ij}^\star.
\end{equation}
The normalized minimum-image Cartesian distance becomes
\begin{equation}
    d^\mathrm{mi}_{ij} = \frac{\|  \Delta\mathbf{f}^\mathrm{mi}_{ij} \mathbf{L}(\mathbf{y})\|_2}{s(\mathbf{y})}
\end{equation}
which is normalized by a cell scale divisor $s(\mathbf{y}) = (a + b + c) / 3$ derived from the lattice vector lengths.

The pairwise geometry is featurized as
\begin{equation}
\mathbf{e}_{ij}
=
\left[
\mathrm{SinEmb}(\Delta\mathbf{f}^\mathrm{mi}_{ij};N_{\mathrm{freq}}^{\mathrm{edge}})
\,\middle\|\,
\left\{\exp[-\gamma(d^\mathrm{mi}_{ij}-\mu_q)^2]\right\}_{q=1}^{N_{\mathrm{rbf}}}
\,\middle\|\,
\mathbf{y}^{\mathrm{Y1}}(\mathbf{y})
\right].
\end{equation}
Here, \(N_{\mathrm{freq}}^{\mathrm{edge}}\) and \(N_{\mathrm{rbf}}\) are the numbers of Fourier and radial basis features, respectively; \(\mu_q\) are uniformly spaced radial centers, \(\gamma\) is their inverse squared spacing, and \(\mathbf{y}^{\mathrm{Y1}}(\mathbf{y})\) denotes the Y1 lattice features computed from \(\mathbf{y}\).

The bias for attention head \(h\) is
\begin{equation}
[\mathbf{B}(\mathbf{c})]_{hij}
=
g_h(s)
\left(
-\operatorname{softplus}(w_h)d^\mathrm{mi}_{ij}
+
[E_{\mathrm{edge}}(\mathbf{e}_{ij})]_h
\right),
\end{equation}
where \(w_h\) is a learned head-specific distance parameter, \(E_{\mathrm{edge}}\) is a two-layer MLP with hidden dimension \(d_{\mathrm{edge}}\), and
\begin{equation}
g_h(s)
=
\operatorname{sigmoid}
\left(
\operatorname{softplus}(\alpha_h)
[-\log(\max(s,\epsilon))]
+\beta_h
\right).
\end{equation}
Here, \(s\) is the left-foot time supplied to the DiT/SiT backbone, \(\alpha_h\) and \(\beta_h\) are learned head-specific parameters, and \(\epsilon>0\) is a numerical-stability constant. The diagonal of the edge bias and all entries involving padded atoms or the lattice token are set to zero.

The geometry bias is added directly to the scaled dot-product attention logits. For attention head \(h\),
\begin{equation}
\mathrm{Attn}_h
=
\operatorname{softmax}
\left(
\frac{\mathbf{Q}_h\mathbf{K}_h^\top}{\sqrt{d_h}}
+
\mathbf{B}_h(\mathbf{c})
+
\mathbf{M}
\right)\mathbf{V}_h,
\end{equation}
where \(\mathbf{Q}_h\), \(\mathbf{K}_h\), and \(\mathbf{V}_h\) are the query, key, and value projections, \(d_h\) is the head dimension, and \(\mathbf{M}\) is the key-padding mask, taking value \(-\infty\) for padded key tokens and zero otherwise. The remainder of the architecture follows standard SiT/DiT token-based setups, using adaLN-conditioned self-attention and MLP residual blocks.

For predicting endpoints which are periodic in the domain $[0,1)$ we parameterize the endpoint prediction of denoised fractional coordinates, 
\begin{equation}
    \hat{\mathbf{F}}_1 = \frac{\mathrm{atan2}(\hat{\mathbf{u}}_1, \hat{\mathbf{v}}_1)}{2\pi} \,\, \texttt{mod} \,\, 1.0,
\end{equation}
Here, $\hat{\mathbf{u}}_1$ and $\hat{\mathbf{v}}_1$ are predicted by the fractional-coordinate head. We use this so-called circular prediction mode for all realizations of OMatG-flash. $\hat{\mathbf{A}}_1$ and $\hat{\mathbf{y}}_1$ are predicted by linear layer output heads.

\paragraph{Data Augmentation} OMatG-flash is not rotation-equivariant nor is it translation invariant however, since $\mathbf{F}$ and $\mathbf{A}$ are naturally rotation invariant, we need only the lattice in handling rotations. Our choice of modeling $\mathbf{y}$ is rotation-invariant and, thus, we do not consider data augmentation for global rotations of the crystal. OMatG-flash is, however, not translation-invariant and $\mathbf{F}$ is sensitive to this. We therefore augment the data with random translations in 3D during training.
\label{app:data_augment}

\section{Hyperparameter Selection}

\begin{table}[H]
\centering
\small
\caption{\textbf{OMatG-flash hyperparameters.}}
\label{tab:hyperparameters}
\renewcommand{\arraystretch}{1.08}
\begin{tabularx}{\linewidth}{@{}Xr@{}}
\toprule
\textbf{Hyperparameter} & \textbf{Value} \\
\midrule
\multicolumn{2}{@{}l}{\textbf{Architecture}} \\
\addlinespace[2pt]
Atomic descriptor dimension $d_A$ & 24 \\
Supported atomic numbers $Z$ & $1$--$100$ \\
Transformer width $d$ & 512 \\
Transformer layers & 18 \\
Attention heads & 16 \\
Coordinate Fourier frequencies & 32 \\
Time-embedding dimension & 256 \\
\midrule
\multicolumn{2}{@{}l}{\textbf{Flow map objective}} \\
\addlinespace[2pt]
Atomic weights $(\lambda_{\mathrm{T},A},\lambda_{\mathrm{C},A})$ & $(2.5,2.5)$ \\
Coordinate weights $(\lambda_{\mathrm{T},F},\lambda_{\mathrm{C},F})$ & $(1.0,1.0)$ \\
Lattice weights $(\lambda_{\mathrm{T},y},\lambda_{\mathrm{C},y})$ & $(0.05,0.05)$ \\
\midrule
\multicolumn{2}{@{}l}{\textbf{Optimization}} \\
\addlinespace[2pt]
Batch size & 256 \\
Optimizer & AdamW \\
Learning rate & $5\times10^{-4}$ \\
AdamW coefficients $(\beta_1,\beta_2)$ & $(0.9,0.999)$ \\
AdamW $\epsilon_{\mathrm{opt}}$ & $10^{-8}$ \\
Weight decay & $10^{-2}$ \\
EMA decay & 0.999 \\
Gradient clipping threshold & 1.0 \\
\midrule
\multicolumn{2}{@{}l}
{\textbf{RAM post-training}} \\
\addlinespace[2pt]
Atomic weights ($\lambda_{\mathrm{RAM}, A}, \lambda_{\mathrm{C}, A}$) & 
($-, -$) \\
Coordinate weights ($\lambda_{\mathrm{RAM}, F}, \lambda_{\mathrm{C}, F}$) & 
(0.520, 0.002) \\
Lattice weights ($\lambda_{\mathrm{RAM}, y}, \lambda_{\mathrm{C}, y}$) & 
(0.074, 0.002) \\
RAM energy weight $\lambda_E$ & 
0.2 \\
RAM number groups & 16\\ RAM group size & 64 \\
RAM noising replicas per sample & 4 \\
RAM learning rate & $1.0 \times 10^{-4}$ \\
RAM training steps & 250 \\
\bottomrule
\end{tabularx}
\end{table}

OMatG-flash's hyperparameters are summarized in Table~\ref{tab:hyperparameters}. We minimize flow map and RAM losses per-field using the weights specified. The channel-specific tangency and consistency weights are normalized internally, so only their relative values matter. Hyperparameters for RAM were fine-tuned using Ray Tune~\cite{liaw2018tune}; initialization for pretraining was derived from intuition. We use a standard normal base distribution which, after wrapping the fractional coordinate onto the flat torus, is effectively uniform. At inference time we use a uniform time grid.

\section{Reinforce Adjoint Matching}
\label{app:ram}

Reinforce adjoint matching frames fine-tuning an ODE velocity field in terms of stochastic optimal control. The central objective is to draw samples from a tilted distribution
\begin{equation}
    \rho^\star(x) \propto \rho(x)e^{r(x)}
\label{eq:tilted_rho}
\end{equation}
where $r$ is a scalar-valued reward function \cite{potaptchik_meta_2026}. In the stochastic dynamical transport setting, samples from $\rho^\star$ can be drawn by sampling random Gaussian noise $x_0$ and integrating an SDE similar to Equation~\eqref{eq:reverse_diffusion_sde} but written forwards in time
\begin{equation}
	dx_t = \left[ f_t(x_t) + \sigma_t^2 \nabla_{x_t} \log \rho_t(x_t) + \sigma^2_t u^\star_t(x_t)  \right] \mathrm{d}t + \sigma_t \mathrm{d} \overrightarrow{W}_t.
\end{equation}
where $u^\star_t$ is a control field which, to sample from Equation~\eqref{eq:tilted_rho}, should be chosen to maximize a terminal reward regularized by a path cost term
\begin{equation}
    u^\star = \operatorname*{arg\,max}_u \mathbb{E} \left[ r(x_1) - \frac{1}{2} \int_0^1 \sigma_t^2 \| u_\tau(x_\tau) \|^2 \, \mathrm{d} \tau \right].
\end{equation}

We write the optimal controller as $u^\star_t(x) = \nabla_x V_t(x)$ where $V_t$ is the value function \cite{domingo-enrich_adjoint_2025}
\begin{equation}
    \nabla_x V_t(x) = \mathbb{E} \left[ r(x_1) \nabla_{x_t} \log \rho_{1|t} (x_1^{u^\star} | x_t) \big| x_t = x \right] - \frac{1}{2} \mathbb{E} \left[ \nabla \int_t^1 \sigma_t^2 \| u_\tau^\star(x_\tau) \|^2 \, \mathrm{d}\tau \bigg| x_t = x \right].
\end{equation}
Dropping the path cost term, we arrive at an approximate value function gradient used in RAM:
\begin{equation}
    \nabla_x V_t(x) \approx \mathbb{E} \left[ r(x_1) \nabla_{x_t} \log \rho_{1|t} (x_1^{u^\star} | x_t) \big| x_t = x \right].
\label{eq:value_function_grad}
\end{equation}
The superscript $u^\star$ indicates that the sample was generated using the controller.

\paragraph{RAM Fixed Point Objective} In order to estimate the control in Equation~\eqref{eq:value_function_grad} we recall the uncontrolled SDE drift
\begin{equation}
    b_t(x) = v_t(x) + \frac{\sigma_t^2}{2} \nabla_x \log \rho_t(x),
\end{equation}
and controlled SDE drifts
\begin{equation}
    b_t(x) + \sigma^2_t u_t^\star(x) = v_t^\star(x) + \frac{\sigma_t^2}{2} \nabla_x \log \rho_t^\star(x),
\end{equation}
Subtracting gives
\begin{equation}
    \sigma_t^2 u_t^\star(x) = \left(v_t^\star(x) - v_t(x)\right) + \frac{\sigma_t^2}{2}\left(\nabla_x \log \rho^\star_t(x) - \nabla_x \log \rho_t(x)\right).
\end{equation} 
Provided that $\rho_0$ is Gaussian and that we use a memoryless schedule, we can write $\nabla_x \log \rho_t(x) = 2 \sigma_t^{-2}(v_t(x) - \dot{\beta}_t \beta^{-1}_t x)$. Using this score identity, one can show that $v_t^\star$ satisfies a fixed point
\begin{align}
    \sigma_t^2 u_t^\star(x) &= 2 (v_t^\star(x) - v_t(x)) \\
    v_t^\star(x) &\approx v_t(x) + \frac{\sigma_t^2}{2}\mathbb{E} \left[ r(x_1) \nabla_{x_t} \log \rho_{1|t} (x_1^{u^\star} | x_t) \big| x_t = x \right]
\end{align}
The last step is to invoke the Bayes bridge score identity shown in the RAM derivation for the linear interpolant schedule ($\alpha_t = 1-t, \beta_t = t$)\ \cite{bergmeister_reinforce_2026},
\begin{equation}
    \nabla_{x_t} \log \rho_{1|t}(x_1 | x_t) = \frac{t}{1-t}((x_1 - x_0) - v_t^\star(x_t)),
\end{equation}
and to enforce $\sigma^2_t/2 = (1-t)/t$ giving the RAM fixed point condition\footnote{We emphasize that, in the original RAM literature, the problem is framed in reverse time where $\rho_0$ is clean data and $\rho_1$ is noise.
Thus, many of the relations in their text have been adjusted to reflect our time convention.}.
\begin{equation}
    v_t^\star(x) - v_t(x) = \mathbb{E} \left[ r(x_1) ( \dot{x}_t - v_t^\star(x_t)) | x_t = x \right].
\end{equation}

\begin{figure}[t]
\centering
\includegraphics[width=1.0\linewidth]{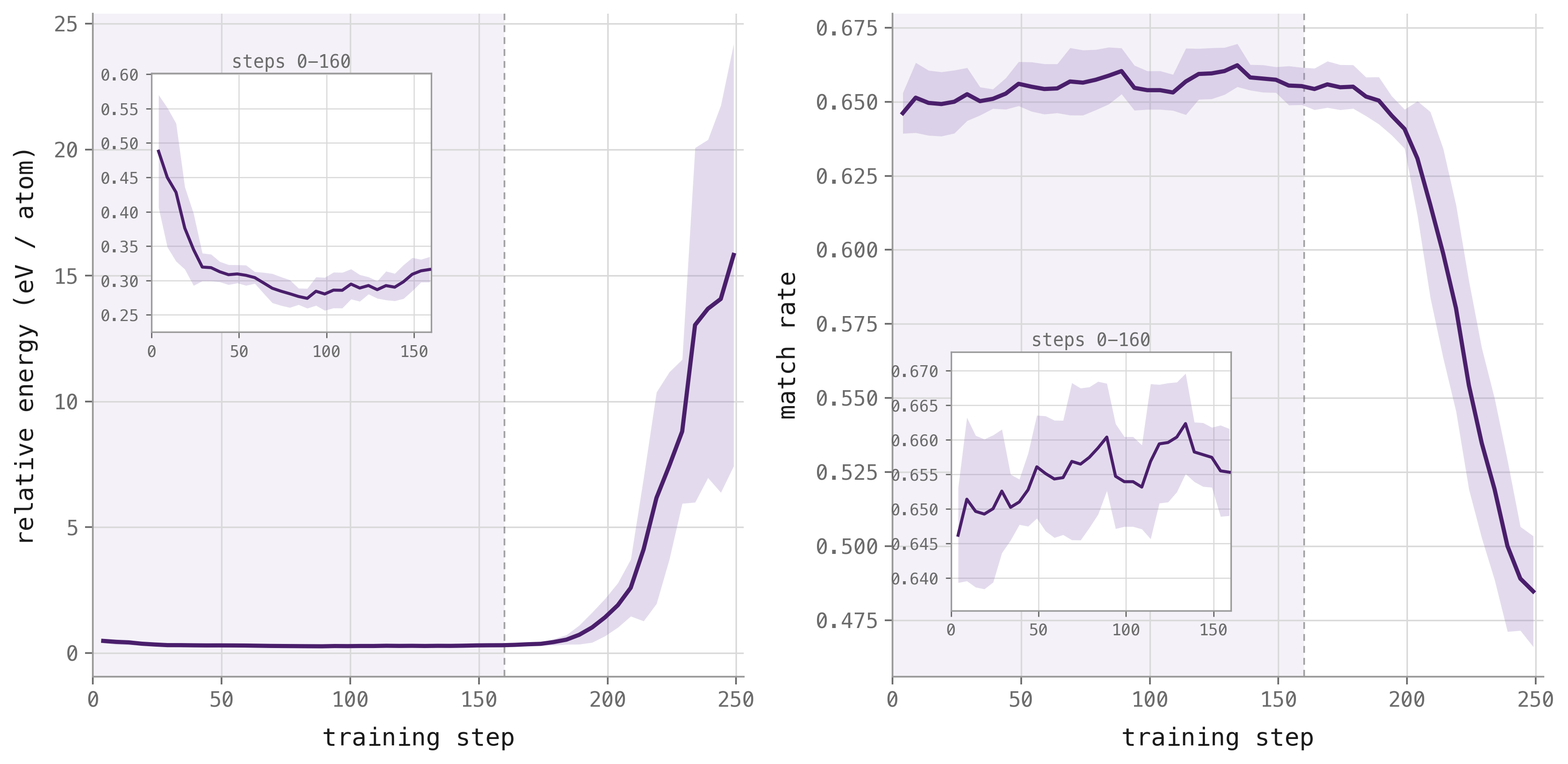}
\caption{\textbf{Instability in RAM Minimization} \, Early during post-training optimization with no relaxation applied the energy of generated samples decreases by over 0.2 eV/atom on average as shown in the inset. The one-to-one match rate trends slightly upward. Later, the RAM and consistency losses become unstable and model performance collapses.}
\label{fig:ram_instability}
\end{figure}

\paragraph{Application to OMatG-flash}
We highlight that the application of RAM to our OMatG-flash flow map improves performance notably (Section~\ref{sec:experiments}).
We do note, however, that in the limit of long training times, we encounter instability and reward collapse and we illustrate this effect in Figure~\ref{fig:ram_instability}. 
Applying relaxation steps to the generated structure before reward computation improved stability of the algorithm.
The numbers for OMatG-flash-RAM reported in Tables \ref{tab:mr_rmse} and \ref{tab:metre_mp20ps} are obtained using the checkpoint which exhibited the lowest average relative energy during training with relaxation applied before reward computation. 

We do note that on $\mathbb{T}^{N \times 3}_{[0,1)}$ the application of the velocity-score reparameterization given in Equation~\eqref{eq:reparam} is heuristic and we do not claim an extension of RAM to arbitrary manifolds is theoretically rigorous nor efficacious. Rewards are normalized to group-relative advantages following \citep{bergmeister_reinforce_2026}.

\section{Benchmark Metrics}
\label{app:lemat_metrics}

\subsection{Crystal Structure Prediction}

\paragraph{Match Rate and RMSE}
Match rate and RMSE are two canonical methods for measuring the performance of generative models on the CSP task. 
Both rely on the pymatgen \texttt{StructureMatcher} algorithm which determines a match within some specified tolerance by performing translations, rotations, and unimodular lattice operations. 
If a match is found, an RMSE can be computed which is then normalized by $\sqrt[3]{V/N_{\mathrm{atoms}}}$ with $V$ the volume of the cell and $N_\mathrm{atoms}$ the number of atoms.
For CSP metrics we use the settings most often used in the generative modeling literature: \texttt{stol}=0.5, \texttt{ltol}=0.3, \texttt{angle\_tol}=10.0. 
This metric is generally computed one-to-one where each generated structure is compared to its index-matched test set reference.

\begin{figure}[t]
\centering
\includegraphics[width=1.0\linewidth]{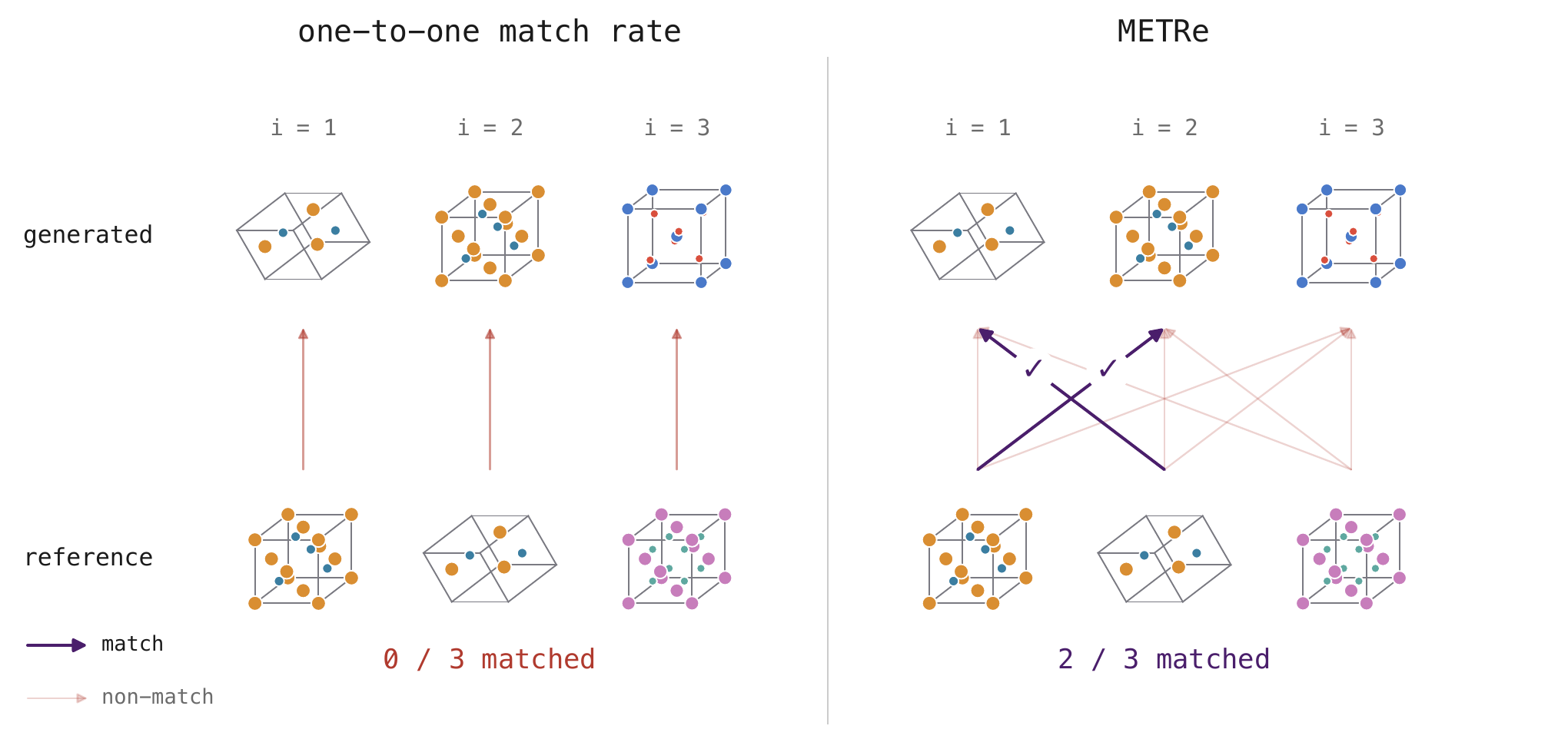}
\caption{\textbf{Concept of METRe vs. One-To-One Matching} \, The one-to-one method (\textbf{Left}) counts three non-matches despite the presence of two pairs of structures that \emph{could} indeed be matched. 
METRe (\textbf{Right}) correctly scans the generated set for each reference structure, determining that two matches are present in the generated set.}
\label{fig:metre_concept}
\end{figure}

\paragraph{METRe and cRMSE}
It has been observed that one-to-one match rate is a suboptimal metric for determining CSP performance \cite{martirossyan_all_2025}. 
This is due to the phenomenon of polymorphism in crystal structures, where a single composition can yield many diverse crystal structures.
METRe attempts to combat this by instead comparing each reference structure against \emph{every} generated structure. 
cRMSE is then a corrected version of RMSE where non-matched structures appear as an explicit penalty in the cRMSE computation
\begin{equation}
    \mathrm{cRMSE}\left(\{\mathbf{c}_{\mathrm{gen},j}\}_{j=1}^{N_\mathrm{gen}}; \{\mathbf{c}_{\mathrm{ref},i}\}_{i=1}^{N_\mathrm{ref}} \right) = \frac{1}{N_\mathrm{ref}} \sum_{i \in \mathcal{X}} \min_{j \in \mathcal{J}_i}\mathrm{RMSE}(\mathbf{c}_{\mathrm{gen}, j}, \mathbf{c}_{\mathrm{ref}, i}) + \texttt{stol} \cdot \frac{\left( N_\mathrm{ref} - N_\mathrm{ref}^\mathrm{match}\right)}{N_\mathrm{ref}} 
\end{equation}
where $\mathbf{c}_\mathrm{gen}$ is a generated crystal, 
$\mathbf{c}_\mathrm{ref}$ is a reference crystal, $N_\mathrm{ref}$ is the number of reference test set structures, and $N_\mathrm{ref}^\mathrm{match}$ is the number of matched structures. We consider $\mathcal{X}$ as
\begin{equation}
    \mathcal{X} = \{ i \in \{ 1, ..., N_\mathrm{ref} \} \,\, : \,\, \mathcal{J}_i \neq \emptyset \}
\end{equation}
where we have $\mathcal{J}_i$ as
\begin{equation}
    \mathcal{J}_i = \{ j \in \{ 1, ..., N_\mathrm{gen} \} \,\, : \,\, \mathbf{c}_{\mathrm{gen}, j} \,\, \mathrm{matches} \,\, \mathbf{c}_{\mathrm{ref}, i} \}.
\end{equation}
We illustrate METRe vs. one-to-one matching in Figure~\ref{fig:metre_concept}. 

\paragraph{Validity} For Table~\ref{tab:mr_rmse} we report results for all structures (\texttt{valid or invalid}) along with values restricted to only those passing certain validity checks (\texttt{valid}). Validity is assessed first via structural assessment followed by SMACT chemical validity. Structural checks ensure that no interparticle distances are less than 0.5\AA, the cell volume is greater than 0.1\AA\textsuperscript{3}, and that a polar sine cutoff
\begin{equation}
    V / (a \times b \times c) \geq 10^{-3}
\end{equation}
 is not violated where $V$ is the volume of the cell and $a,b,c$ are lattice vector lengths. If these are satisfied, a crystal structure must then pass a SMACT compositional validity check \citep{davies_smact_2019} followed by a chemical fingerprint validity check.

\subsection{De Novo Generation}
All DNG benchmark metrics are computed with code provided by \href{https://github.com/LeMaterial/lemat-genbench}{\textcolor{blue}{LeMat-GenBench}} in their open source implementation using the \texttt{scripts/run\_benchmarks.py} script with \texttt{comprehensive\_multi\_mlip\_hull} configuration flag and \texttt{structure-matcher} fingerprinting \cite{betala_lematgenbench_2026}. 
The subsequent results are extracted with \texttt{scripts/extract\_benchmark\_metrics.py}.
We prerelax our 2500 generated structures using MACE for up to 1000 steps with a force tolerance of 0.02 eV/\AA\, before benchmarking with LeMat-GenBench. To expand on our results in Table~\ref{tab:sun} we include raw numbers as reported by LeMat for all DNG metrics in Table~\ref{tab:sun_raw}

\begin{table}[t]
\centering
\caption{\textbf{DNG Raw Numbers.} LeMat-GenBench reports public leaderboard results using one decimal place. For transparency we report the raw numbers computed by the LeMat-GenBench evaluator and extractor scripts. Numbers are computed for an initial pool of 2500 structures. We take numbers for competing models from the public LeMat-GenBench leaderboard.}
\vspace{4pt}
\label{tab:sun_raw}
\begin{tabular}{l ccccc}
\toprule
Method & Valid $\uparrow$ & Novel $\uparrow$ & Stable $\uparrow$ & S.U.N. $\uparrow$ & S.U.N. + M.S.U.N. $\uparrow$ \\
\midrule
DiffCSP     & 2392 & 1654 & 58 & 3 & 215 \\
MatterGen   & 2392 & 1762 & 49 & 6 & 380 \\
OMatG       & 2361 & 1255 & 284 & 24 & 464 \\
MCFlow      & 2429 & 1306 & 298 & 18 & 490 \\
Crystalite  & 2430 & 1331 & 317 & 38 & 604 \\
\midrule
OMatG-flash$_4$ & 2409 & 1672 & 180 & 34 & 455 \\
OMatG-flash$_8$ & 2420 & 1584 & 196 & 26 & 537 \\
OMatG-flash$_{16}$ & 2420 & 1524 & 221 & 29 & 564 \\
\bottomrule
\end{tabular}
\end{table}

\paragraph{Stability}
Stability is computed using a series of foundation model interatomic potentials.
In order to determine if a material is stable or not it is input to UMA \cite{wood_uma_2025}, ORB \cite{neumann_orb_2024}, and MACE \cite{batatia2023foundation}. 
The average energy-above-hull returned by these models is compared with the LeMat convex hull of stable phases corresponding to that reduced composition \cite{bartel_critical_2020}. 
A material is stable if and only if its energy-above-hull, $E_\mathrm{hull} \leq 0.0$ eV/atom. 
Metastability for the purposes of M.S.U.N. computation is similarly defined, triggering if $0.0 < E_\mathrm{hull} \leq 0.1$ eV/atom. 
Additionally, a material can only be considered if at least two of the three MLIPs in the ensemble return an energy for that entry.

\paragraph{Uniqueness and Novelty}
Uniqueness and novelty are computed in the same way but using different reference data. 
For both metrics pymatgen's \texttt{StructureMatcher} algorithm is used.
The tolerances used by LeMat-GenBench are stricter than we use for computing CSP and are set \texttt{stol}=0.3, \texttt{ltol}=0.1, \texttt{angle\_tol}=5.0. 
Uniqueness is determined by comparison to all structures in the generated set. 
Novelty, on the other hand, is determined against a held-out LeMat-Bulk reference set \cite{siron_lematbulk_2025}.

\subsection{Wall-Clock Inference Speed} \label{app:timing}
We measure sampling speed using the conventions established in the literature.
To measure results in accordance with the literature, we use one NVIDIA H100 GPU choosing a batch size of 8192 structures for the DNG task with the hyperparameters as reported in Appendix~\ref{app:OMatG-flash_deets}. 
We use \texttt{bfloat16} inference which is consistent with how OMatG-flash was trained.  We define generation time as the time required to sample $\mathbf{c}_0$, perform inference with OMatG-flash, and decode the data into a standard unit cell representation, excluding I/O.
For reporting the Pareto frontier in Figure~\ref{fig:concept} we use baseline timing measurements as computed by Crystalite \cite{veljković_crystalite_2026}. 
For reporting Crystalite itself we use their ``standard'' inference setting as they indicate that this is the primary timing one should consider. 
We emphasize that, even with their reported optimized inference time of 5.14 seconds per thousand structures, our model at NFE=16 is an order of magnitude faster, recording 0.535 seconds per thousand structures.

\end{document}